\documentclass[manuscript,acmsmall,screen,nonacm]{acmart}

\AtBeginDocument{%
  \providecommand\BibTeX{{%
    \normalfont B\kern-0.5em{\scshape i\kern-0.25em b}\kern-0.8em\TeX}}}

\usepackage{hyperref}       
\usepackage{amsfonts}       
\usepackage{color}
\usepackage{tikz}
\usepackage[edges]{forest}
\definecolor{hiddendraw}{RGB}{205, 44, 36}
\definecolor{hidden-blue}{RGB}{194,232,247}
\definecolor{hidden-orange}{RGB}{243,202,120}
\definecolor{hidden-yellow}{RGB}{242,244,193}
\tikzstyle{mybox}=[
    rectangle,
    draw=hiddendraw,
    rounded corners,
    text opacity=1,
    minimum height=1.5em,
    minimum width=5em,
    inner sep=2pt,
    align=center,
    fill opacity=.5,
    ]

\usepackage{mydef}
\usepackage{amssymb}
\usepackage{pifont}

\usepackage{amsthm}

\usepackage{multirow} 

\usepackage{subfigure}

\theoremstyle{definition}
\newtheorem{definition}{Definition}[section]

\begin{document}

\title{A Survey of Transformers}

\author{Tianyang Lin}
\email{tylin20@fudan.edu.cn}
\orcid{0000-0003-1193-6472}
\author{Yuxin Wang}
\author{Xiangyang Liu}
\author{Xipeng Qiu}
 \authornote{Corresponding Author.}
 \email{xpqiu@fudan.edu.cn}
 \orcid{0000-0001-7163-5247}
\affiliation{%
  \institution{School of Computer Science, Fudan University}
  \city{Shanghai}
  \country{China}
  \postcode{200433}
}
\affiliation{%
  \institution{Shanghai Key Laboratory of Intelligent Information Processing, Fudan University}
  \city{Shanghai}
  \country{China}
  \postcode{200433}
}

\renewcommand{\shortauthors}{Lin et al.}

\begin{abstract}
  Transformers have achieved great success in many artificial intelligence fields, such as natural language processing, computer vision, and audio processing. Therefore, it is natural to attract lots of interest from academic and industry researchers. Up to the present, a great variety of Transformer variants (a.k.a. X-formers) have been proposed, however, a systematic and comprehensive literature review on these Transformer variants is still missing. In this survey, we provide a comprehensive review of various X-formers. We first briefly introduce the vanilla Transformer and then propose a new taxonomy of X-formers. Next, we introduce the various X-formers from three perspectives: architectural modification, pre-training, and applications. Finally, we outline some potential directions for future research.
\end{abstract}



\maketitle

\section{Introduction}\label{sec:intro}

Transformer~\cite{vaswani2017attention} is a prominent deep learning model that has been widely adopted in various fields, such as natural language processing (NLP), computer vision (CV) and speech processing. Transformer was originally proposed as a sequence-to-sequence model~\cite{sutskever14seq2seq} for machine translation. Later works show that Transformer-based pre-trained models (PTMs)~\cite{qiu2020ptms} can achieve \textit{state-of-the-art} performances on various tasks. As a consequence, Transformer has become the go-to architecture in NLP, especially for PTMs. In addition to language related applications, Transformer has also been adopted in CV~\cite{parmar2018imagexformer,carion20detr,dosovitskiy2020vit}, audio processing~\cite{dong18speechxformer,gulati20conformer,chen20streamingxformer} and even other disciplines, such as chemistry~\cite{schwaller19olecularxformer} and life sciences~\cite{rivese21biological}.

Due to the success, a variety of Transformer variants (a.k.a. X-formers) have been proposed over the past few years.
These X-formers improve the vanilla Transformer from different perspectives.
\begin{enumerate}
  \item \textit{Model Efficiency}. A key challenge of applying Transformer is its inefficiency at processing long sequences mainly due to the computation and memory complexity of the self-attention module.
The improvement methods include lightweight attention (e.g. sparse attention variants) and Divide-and-conquer methods (e.g., recurrent and hierarchical mechanism).
  \item \textit{Model Generalization}. Since the transformer is a flexible architecture and makes few assumptions on the structural bias of input data, it is hard to train on small-scale data. The improvement methods include introducing structural bias or regularization, pre-training on large-scale unlabeled data, etc.
  \item \textit{Model Adaptation}. This line of work aims to adapt the Transformer to specific downstream tasks and applications.
\end{enumerate}

In this survey, we aim to provide a comprehensive review of the Transformer and its variants. Although we can organize X-formers on the basis of the perspectives mentioned above, many existing X-formers may address one or several issues. For example, sparse attention variants not only reduce the computational complexity but also introduce structural prior on input data to alleviate the overfitting problem on small datasets. Therefore, it is more methodical to categorize the various existing X-formers and propose a new taxonomy mainly according to their ways to improve the vanilla Transformer: architecture modification, pre-training, and applications.
Considering the audience of this survey may be from different domains, we mainly focus on the general architecture variants and just briefly discuss the specific variants on pre-training and applications.

The rest of the survey is organized as follows. Sec.~\ref{sec:background} introduces the architecture and the key components of Transformer. Sec.~\ref{sec:taxonomy} clarifies the categorization of Transformer variants. Sec.~\ref{sec:attention}$\sim$\ref{sec:other_module} review the module-level modifications, including attention module, position encoding, layer normalization and feed-forward layer. Sec.~\ref{sec:beyond} reviews the architecture-level variants. Sec.~\ref{sec:ptm} introduces some of the representative Transformer-based PTMs. Sec.~\ref{sec:app} introduces the application of Transformer to various different fields. Sec.~\ref{sec:discussion} discusses some aspects of Transformer that researchers might find intriguing and summarizes the paper.

\section{Background}\label{sec:background}
\subsection{Vanilla Transformer}\label{sec:vanilla_xformer}
\begin{figure}[htbp]
    \centering
    \includegraphics[width=0.7\linewidth]{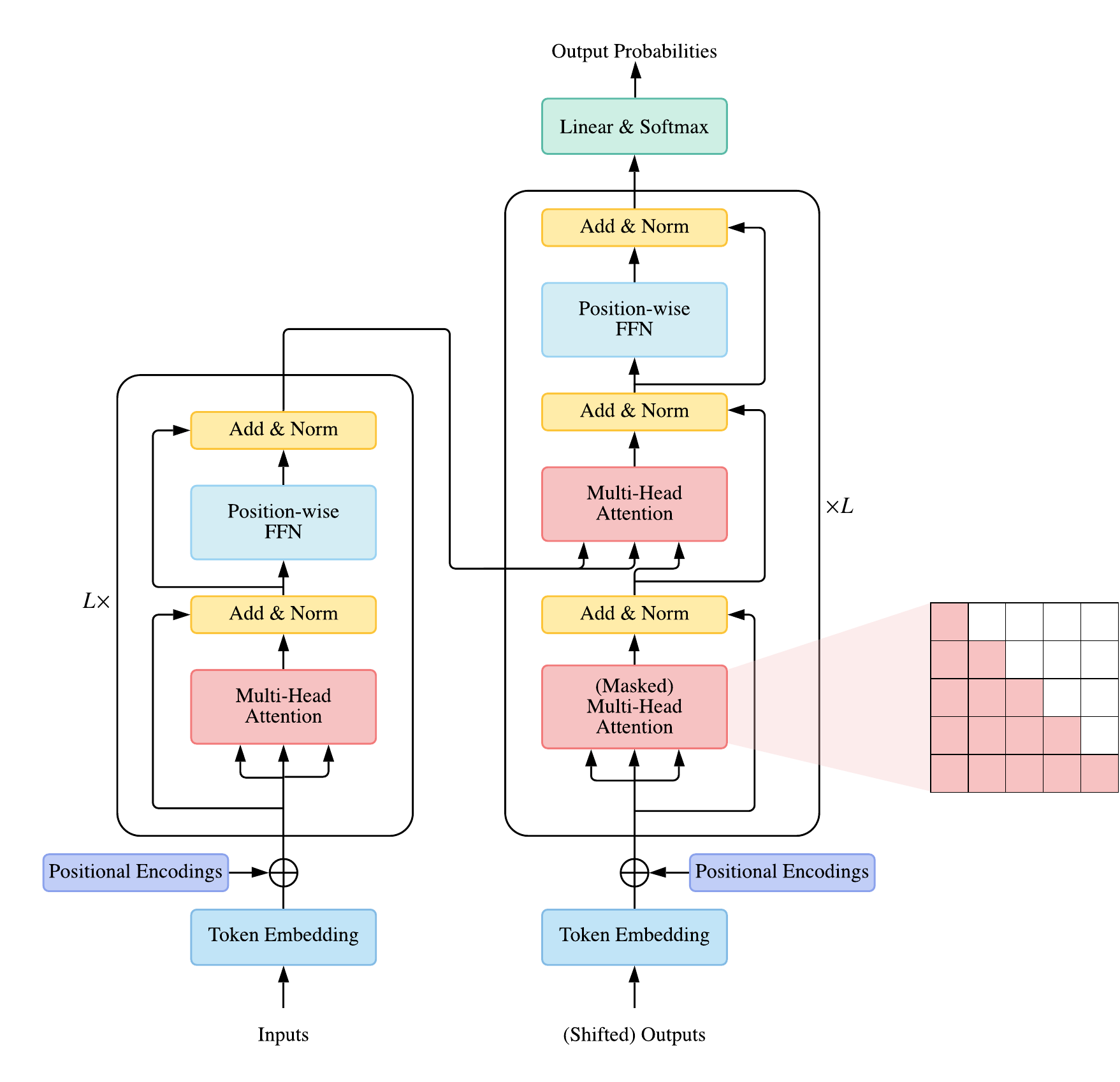}
    \caption{Overview of vanilla Transformer architecture}
    \label{fig:xformer_arch}
\end{figure}

The vanilla Transformer~\cite{vaswani2017attention} is a sequence-to-sequence model and consists of an encoder and a decoder, each of which is a stack of $L$ identical blocks. Each \textit{encoder block} is mainly composed of a multi-head self-attention module and a position-wise feed-forward network (FFN).
For building a deeper model, a residual connection~\cite{he16resnet} is employed around each module, followed by Layer Normalization~\cite{ba16layernorm} module.
Compared to the encoder blocks, decoder blocks additionally insert cross-attention modules between the multi-head self-attention modules and the position-wise FFNs. Furthermore, the self-attention modules in the decoder are adapted to prevent each position from attending to subsequent positions.
The overall architecture of the vanilla Transformer is shown in Fig. \ref{fig:xformer_arch}.

In the following subsection, we shall introduce the key modules of the vanilla Transformer.

\subsubsection{Attention Modules}

Transformer adopts attention mechanism with Query-Key-Value (QKV) model. Given the packed matrix representations of queries $\bQ\in\mathbb{R}^{N\times D_k}$, keys $\bK\in\mathbb{R}^{M\times D_k}$, and values $\bV\in\mathbb{R}^{M\times D_v}$, the scaled dot-product attention used by Transformer is given by\footnote{if not stated otherwise, we use row-major notations throughout this survey (e.g., the $i$-th row in $\bQ$ is the query $\bq_i$) and all the vectors are row vectors by default.}
\begin{equation}\label{attention}
    \mathrm{Attention}(\bQ, \bK, \bV)=\mathrm{softmax}\left(\frac{\bQ\bK^\top}{\sqrt{D_k}}\right)\bV=\bA\bV,
\end{equation}
where $N$ and $M$ denote the lengths of queries and keys (or values); $D_k$ and $D_v$ denote the dimensions of keys (or queries) and values; $\bA=\mathrm{softmax}\left(\frac{\bQ\bK^\top}{\sqrt{D_k}}\right)$ is often called \textit{attention matrix}; softmax is applied in a row-wise manner. The dot-products of queries and keys are divided by $\sqrt{D_k}$ to alleviate gradient vanishing problem of the softmax function.

Instead of simply applying a single attention function, Transformer uses multi-head attention, where the $D_m$-dimensional original queries, keys and values are projected into $D_k$, $D_k$ and $D_v$ dimensions, respectively, with $H$ different sets of learned projections. For each of the projected queries, keys and values, and output is computed with attention according to Eq. \eqref{attention}. The model then concatenates all the outputs and projects them back to a $D_m$-dimensional representation.
\begin{align}
    \mathrm{MultiHeadAttn}(\bQ,\bK,\bV)&=\mathrm{Concat}(\mathrm{head}_1,\cdots,\mathrm{head}_H)\bW^O\label{eq:multihead},\\
    \mathrm{where}\ \mathrm{head}_i&=\mathrm{Attention}(\bQ \bW_i^Q,\bK \bW_i^K, \bV \bW_i^V)\label{eq:headi}.
\end{align}

In Transformer, there are three types of attention in terms of the source of queries and key-value pairs:
\begin{itemize}
    \item \textit{Self-attention}. In Transformer encoder, we set $\bQ=\bK=\bV=\bX$ in Eq. \eqref{eq:multihead}, where $\bX$ is the outputs of the previous layer.
    \item \textit{Masked Self-attention}. In the Transformer decoder, the self-attention is restricted such that queries at each position can only attend to all key-value pairs up to and including that position. To enable parallel training, this is typically done by applying a mask function to the unnormalized attention matrix $\hat{\bA}=\exp(\frac{\bQ\bK^\top}{\sqrt{D_k}})$, where the illegal positions are masked out by setting $\hat{A}_{ij}=-\infty \textrm{ if } i<j$. This kind of self-attention is often referred to as autoregressive or causal attention\footnote{This term seems to be borrowed from the \textit{causal system}, where the output depends on past and current inputs but not future inputs.}.
    \item \textit{Cross-attention}. The queries are projected from the outputs of the previous (decoder) layer, whereas the keys and values are projected using the outputs of the encoder.
\end{itemize}

\subsubsection{Position-wise FFN}
The position-wise FFN\footnote{The parameters are shared across different positions, thus the position-wise FFN can also be understood as two convolution layers with kernel size of 1.} is a fully connected feed-forward module that operates separately and identically on each position
\begin{equation}
    \mathrm{FFN}(\bH') = \mathrm{ReLU}(\bH'\bW^1 + \bb^1)\bW^2+\bb^2,
\end{equation}
where $\bH'$ is the outputs of previous layer, and $\bW^1\in\mathbb{R}^{D_m\times D_f},\bW^2\in\mathbb{R}^{D_f\times D_m},\bb^1\in\mathbb{R}^{D_f},\bb^2\in\mathbb{R}^{D_m}$ are trainable parameters. Typically the intermediate dimension $D_{f}$ of the FFN is set to be larger than $D_m$.

\subsubsection{Residual Connection and Normalization}

In order to build a deep model, Transformer employs a residual connection~\cite{he16resnet} around each module, followed by Layer Normalization~\cite{ba16layernorm}. For instance, each Transformer encoder block may be written as
\begin{align}
    \bH'&=\mathrm{LayerNorm}(\mathrm{SelfAttention}(\bX)+\bX)\\
    \bH&=\mathrm{LayerNorm}(\mathrm{FFN}(\bH')+\bH'),
\end{align}
where $\mathrm{SelfAttention}(\cdot)$ denotes self attention module and $\mathrm{LayerNorm}(\cdot)$ denotes the layer normalization operation.

\subsubsection{Position Encodings}
Since Transformer doesn't introduce recurrence or convolution, it is ignorant of positional information (especially for the encoder). Thus additional positional representation (Detailed discussion in Sec.~\ref{sec:pos_rep}) is needed to model the ordering of tokens.
\subsection{Model Usage}

Generally, the Transformer architecture can be used in three different ways:
\begin{itemize}
    \item \textit{Encoder-Decoder}. The full Transformer architecture as introduced in Sec.~\ref{sec:vanilla_xformer} is used. This is typically used in sequence-to-sequence modeling (e.g., neural machine translation).
    \item \textit{Encoder only}. Only the encoder is used and the outputs of the encoder are utilized as a representation for the input sequence. This is usually used for classification or sequence labeling problems.
    \item \textit{Decoder only}. Only the decoder is used, where the encoder-decoder cross-attention module is also removed. This is typically used for sequence generation, such as language modeling.
\end{itemize}

\subsection{Model Analysis}\label{sec:model-analysis}

To illustrate the computation time and parameter requirements of the Transformer, we analyze the two core components of the Transformer (i.e., the self-attention module and the position-wise FFN) in Table \ref{tab:complexity_attn_ffn}. We assume that the hidden dimension $D_m$ of the model is $D$, and that the input sequence length is $T$. The intermediate dimension of FFN is set to $4D$ and the dimension of keys and values are set to $D/H$ as in \citet{vaswani2017attention}.

\begin{table}[htbp]
    \caption{Complexity and parameter counts of self-attention and position-wise FFN}
    \label{tab:complexity_attn_ffn}
    \centering
    \begin{tabular}{c|c|c}
    \hline
     Module & Complexity & \#Parameters\\
    \hline
     self-attention & $\mathcal O(T^2\cdot D)$ & $4D^2$\\
    \hline
     position-wise FFN & $\mathcal O(T\cdot D^2)$ & $8D^2$\\
    \hline
    \end{tabular}
\end{table}

When the input sequences are short, the hidden dimension $D$ dominates the complexity of self-attention and position-wise FFN. The bottleneck of Transformer thus lies in FFN. However, as the input sequences grow longer, the sequence length $T$ gradually dominates the complexity of these modules, in which case self-attention becomes the bottleneck of Transformer. Furthermore, the computation of self-attention requires that a $T\times T$ attention distribution matrix is stored, which makes the computation of Transformer infeasible for long-sequence scenarios (e.g., long text documents and pixel-level modeling of high-resolution images). One shall see that the goal of increasing the efficiency of Transformer generally leads to the long-sequence compatibility of self-attention, as well as the computation and parameter efficiency of position-wise FFN for ordinary settings.

\subsection{Comparing Transformer to Other Network Types}

\subsubsection{Analysis of Self-Attention}
As a central piece of Transformer, self-attention comes with a flexible mechanism to deal with variable-length inputs. It can be understood as a fully connected layer where the weights are dynamically generated from pairwise relations from inputs. Table \ref{tab:op_complexities} compares the complexity, sequential operations, and maximum path length\footnote{The maximum length of the paths forward and backward signals have to traverse to get from any input position to arbitrary output position. Shorter length implies a better potential for learning long-range dependencies.} of self-attention with three commonly used layer types. We summarize the advantages of self-attention as follows:
\begin{enumerate}
    \item It has the same maximum path length as fully connected layers, making it suitable for long-range dependencies modeling. Compared to fully connected layers, it is more parameter-efficient and more flexible in handling variable-length inputs.
    \item Due to the limited receptive field of convolutional layers, one typically needs to stack a deep network to have a global receptive field. On the other hand, the constant maximum path length enables self-attention to model long-range dependencies with a constant number of layers.
    \item The constant sequential operations and maximum path length make self-attention more parallelizable and better at long-range modeling than recurrent layers.
\end{enumerate}

\begin{table}[ht]
    \caption{Per-layer complexity, minimum number of sequential operations and maximum path lengths for different layer types. $T$ is the sequence length, $D$ is the representation dimension and $K$ is the kernel size of convolutions~\cite{vaswani2017attention}.}
\label{tab:op_complexities}
\begin{center}
\vspace{-1mm}
\begin{tabular}{lccc}
\toprule
Layer Type & Complexity & Sequential & Maximum Path Length  \\
           &per Layer         & Operations &   \\
\hline
\rule{0pt}{2.0ex}Self-Attention & $\mathcal O(T^2 \cdot D)$ & $\mathcal O(1)$ & $\mathcal O(1)$ \\
Fully Connected & $\mathcal O(T^2 \cdot D^2)$ & $\mathcal O(1)$ & $\mathcal O(1)$ \\
Convolutional & $\mathcal O(K \cdot T \cdot D^2)$ & $\mathcal O(1)$ & $\mathcal O(\log_K(T))$ \\
Recurrent & $\mathcal O(T \cdot D^2)$ & $\mathcal O(T)$ & $\mathcal O(T)$ \\
\bottomrule
\end{tabular}
\end{center}
\end{table}

\subsubsection{In Terms of Inductive Bias}
Transformer is often compared against convolutional and recurrent networks. Convolutional networks are known to impose the inductive biases of translation invariance and locality with shared local kernel functions. Similarly, recurrent networks carry the inductive biases of temporal invariance and locality via their Markovian structure~\cite{battaglia2018relational}. On the other hand, the Transformer architecture makes few assumptions about structural information of data. This makes Transformer a universal and flexible architecture. As a side effect, the lack of structural bias makes Transformer prone to overfitting for small-scale data.

Another closely related network type is Graph Neural Networks (GNNs) with message passing~\cite{wu21surveyonGNN}. Transformer can be viewed as a GNN defined over a complete directed graph (with self-loop) where each input is a node in the graph. The key difference between Transformer and GNNs is that Transformer introduces no prior knowledge over how input data are structured — the message passing process in Transformer solely depends on similarity measures over the content.

\section{Taxonomy of Transformers}
\label{sec:taxonomy}

\begin{figure}[htbp]
    \centering
    \includegraphics[width=1.0\linewidth]{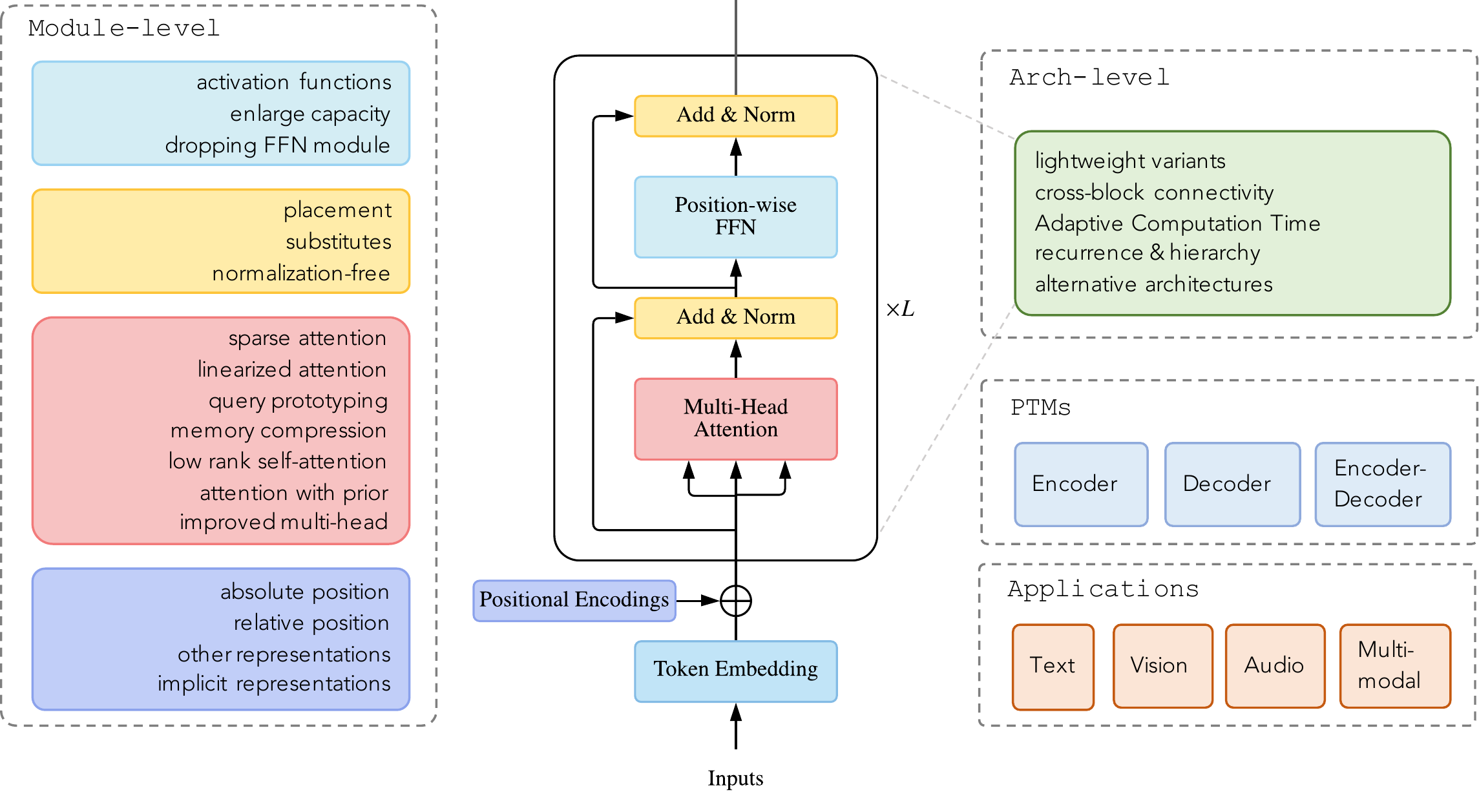}
    \caption{Categorization of Transformer variants.}
    \label{fig:xformer_taxonomy}
\end{figure}

A wide variety of models have been proposed so far based on the vanilla Transformer from three perspectives: types of architecture modification, pre-training methods, and applications.
Fig. \ref{fig:xformer_taxonomy} gives an illustrations of our categorization of Transformer variants.

Fig. \ref{taxonomy_of_xformer} illustrates our taxonomy and some representative models.

\tikzstyle{leaf}=[mybox,minimum height=1em,
fill=hidden-orange!40, text width=20em,  text=black,align=left,font=\tiny,
inner xsep=2pt,
inner ysep=1pt,
]

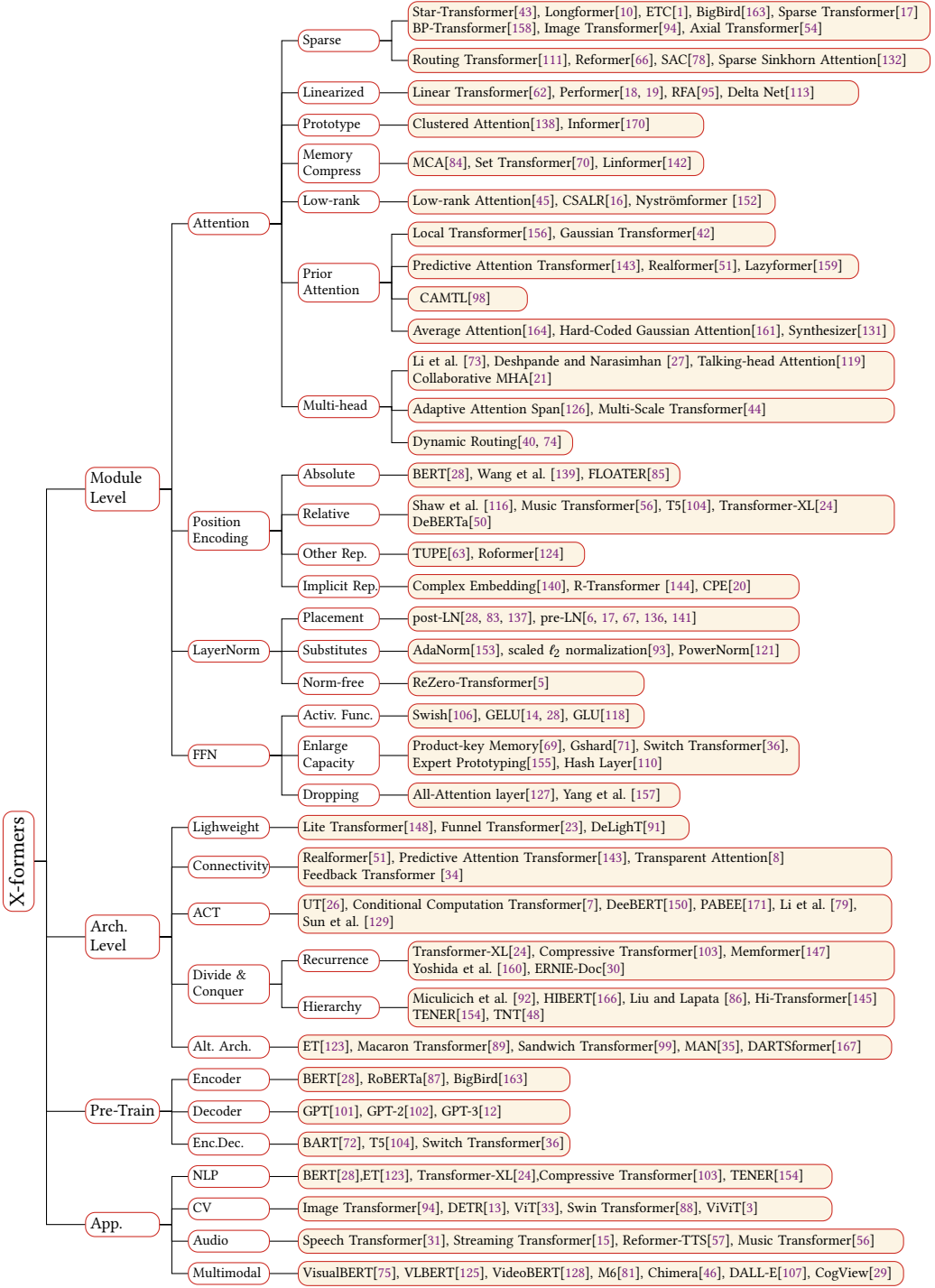
\begin{figure*}[tp]
  \centering
\begin{forest}
  forked edges,
  for tree={
  grow=east,
  reversed=true,
  anchor=base west,
  parent anchor=east,
  child anchor=west,
  base=left,
  font=\small,
  rectangle,
  draw=hiddendraw,
  rounded corners,align=left,
  minimum width=2.5em,
s sep=3pt,
inner xsep=2pt,
inner ysep=1pt,
ver/.style={rotate=90, child anchor=north, parent anchor=south, anchor=center},
  },
  where level=1{text width=2.7em,font=\scriptsize,}{},
  where level=2{text width=3em,font=\tiny}{},
  where level=3{text width=3em,font=\tiny}{},
  [X-formers, ver
    [Module\\Level
        [Attention
           [Sparse
            [Star-Transformer\cite{guo2019startransformer}{,} Longformer\cite{beltagy2020longformer}{,} ETC\cite{ainslie2020etc}{,} BigBird\cite{zaheer2020big}{,} Sparse Transformer\cite{child2019generating}\\
            BP-Transformer\cite{ye2019bptransformer}{,} Image Transformer\cite{parmar2018imagexformer}{,} Axial Transformer\cite{ho19axialxformer}
            ,leaf,text width=21.5em]
            [Routing Transformer\cite{roy2020efficient}{,} Reformer\cite{kitaev2020reformer}{,} SAC\cite{li2020sac}{,} Sparse Sinkhorn Attention\cite{tay2020sparse}
            ,leaf,text width=21.5em]
           ]
           [Linearized
            [Linear Transformer\cite{katharopoulos2020linearxformer}{,} Performer\cite{choromanski2020masked,choromanski2020rethinking}{,} RFA\cite{peng2021random}{,} Delta Net\cite{schlag21fastweight}
            ,leaf,text width=18.5em]
           ]
           [Prototype
            [Clustered Attention\cite{vyas2020clusteredattn}{,} Informer\cite{zhou20informer}
            ,leaf,text width=12em]
           ]
           [Memory\\Compress
            [MCA\cite{liu2018generating}{,} Set Transformer\cite{lee2019set}{,} Linformer\cite{wang2020linformer}
            ,leaf,text width=12em]
           ]
           [Low-rank
            [Low-rank Attention\cite{guo19lowrank}{,} CSALR\cite{chen20compressedlowrank}{,} Nystr{\"{o}}mformer~\cite{xiong21nystromformer}
            ,leaf,text width=15em]
           ]
           [Prior\\Attention
            [Local Transformer\cite{yang18localness}{,} Gaussian Transformer\cite{guo19gaussian}
            ,leaf,text width=15em]
            [Predictive Attention Transformer\cite{wang2021predictive}{,} Realformer\cite{he2020realformer}{,} Lazyformer\cite{ying21lazyformer}
            ,leaf,text width=18.5em]
            [CAMTL\cite{pilault2021conditionally}
            ,leaf,text width=4em]
            [Average Attention\cite{zhang-etal-2018-average}{,} Hard-Coded Gaussian Attention\cite{you2020hardcoded}{,} Synthesizer\cite{synthesizer}
            ,leaf]
           ]
           [Multi-head
            [\citet{li18disagreementregu}{,} \citet{deshpande20guidingattention}{,} Talking-head Attention\cite{shazeer20talkinghead}\\
            Collaborative MHA\cite{cordonnier20collaborate}
            ,leaf,text width=20em]
            [Adaptive Attention Span\cite{Sukhbaatar19adaptivespan}{,} Multi-Scale Transformer\cite{guo2019multiscale}
            ,leaf,text width=20em]
            [Dynamic Routing\cite{li-etal-2019-multiheadrout,gu19capsule}
            ,leaf,text width=6.5em]
           ]
        ]
        [Position\\ Encoding
            [Absolute
                [BERT\cite{devlin2019bert}{,} \citet{wang2021peonbert}{,} FLOATER\cite{liu2020floater}
                ,leaf,text width=11em]
            ]
            [Relative
                 [\citet{shaw2018relative}{,} Music Transformer\cite{huang2018music}{,} T5\cite{raffel2020t5}{,} Transformer-XL\cite{dai2019transformerxl}\\ DeBERTa\cite{he2020deberta}
                ,leaf,text width=20em]
            ]
            [Other Rep.
                [TUPE\cite{ke2020tupe}{,} Roformer\cite{su2021roformer}
                ,leaf,text width=7em]
            ]
            [Implicit Rep.
                [Complex Embedding\cite{wang2020complex}{,} R-Transformer \cite{wang19rxformer}{,} CPE\cite{chu2021conditional}
                ,leaf,text width=16em]
            ]
        ]
        [LayerNorm
            [Placement
                [post-LN\cite{vaswani2017attention,devlin2019bert,liu20understanding}{,} pre-LN\cite{vaswani-etal-2018-tensor2tensor,klein-etal-2017-opennmt,baevski2019adaptive,child2019generating,wang19learningdeep}
                ,leaf,text width=16em]
            ]
            [Substitutes
                [AdaNorm\cite{xu19understanding}{,} scaled $\ell_2$ normalization\cite{nguyen19xformerwotears}{,} PowerNorm\cite{shen2020powernorm}
                ,leaf,text width=16em]
            ]
            [Norm-free
                 [ReZero-Transformer\cite{bachlechner20rezero}
                ,leaf,text width=9.5em]
            ]
        ]
        [FFN
            [Activ. Func.
                [Swish\cite{ramachandran18searching}{,} GELU\cite{chen20igpt,devlin2019bert}{,} GLU\cite{shazeer2020glu}
                ,leaf,text width=9.5em]
            ]
            [Enlarge\\Capacity
                [Product-key Memory\cite{lample2019large}{,} Gshard\cite{lepikhin20gshard}{,} Switch Transformer\cite{fedus2021switch}{,}\\ Expert Prototyping\cite{yang2021exploring}{,} Hash Layer\cite{roller2021hash}
                ,leaf,text width=16em]
            ]
            [Dropping
                [All-Attention layer\cite{sukhbaatar2019augmenting}{,} \citet{yang20xformerdec}
                ,leaf,text width=11em]
            ]
        ]
    ]
    [Arch.\\Level
        [Lighweight
            [Lite Transformer\cite{wu20longshortrange}{,} Funnel Transformer\cite{dai20funneltransformer}{,} DeLighT\cite{mehta2020delight}
            ,leaf,text width=16em]
        ]
        [Connectivity
            [Realformer\cite{he2020realformer}{,} Predictive Attention Transformer\cite{wang2021predictive}{,} Transparent Attention\cite{Chen18transparentxformer}\\
             Feedback Transformer~\cite{fan2021feedbackmem}
            ,leaf,text width=24.5em]
        ]
        [ACT
            [UT\cite{universalxformer}{,} Conditional Computation Transformer\cite{CCTransformer}{,} DeeBERT\cite{xin-etal-2020-deebert}{,} PABEE\cite{zhou2020bert}{,} \citet{li21accelerating}{,} \\\citet{sun2021early}
            ,leaf,text width=24.5em]
        ]
        [Divide \& \\
        Conquer
            [Recurrence
                [Transformer-XL\cite{dai2019transformerxl}{,} Compressive Transformer\cite{rae2019compressive}{,} Memformer\cite{wu2020memformer}\\ \citet{yoshida20recurrenceptm}{,} ERNIE-Doc\cite{ding2020erniedoc}
            ,leaf,text width=20em]
            ]
            [Hierarchy
                [\citet{miculicich-etal-2018-document}{,} HIBERT\cite{zhang-etal-2019-hibert}{,} \citet{liu-lapata-2019-hierarchical}{,} Hi-Transformer\cite{wu2021hitransformer}\\ TENER\cite{yan2019tener}{,} TNT\cite{han2021tnt}
                ,leaf,text width=20em]
            ]
        ]
        [Alt. Arch.
             [ET\cite{so19evolvedxformer}{,} Macaron Transformer\cite{lu19macaron}{,} Sandwich Transformer\cite{ofir20sandwichxformer}{,} MAN\cite{fan-etal-2021-mask}{,} DARTSformer\cite{zhao2021dartsformer}
            ,leaf,text width=24.5em]
        ]
    ]
    [Pre-Train
        [Encoder
            [BERT\cite{devlin2019bert}{,} RoBERTa\cite{liu2019roberta}{,} BigBird\cite{zaheer2020big},leaf,text width=11em]
        ]
        [Decoder
            [GPT\cite{radford2018gpt}{,} GPT-2\cite{radford2019gpt2}{,} GPT-3\cite{brown20gpt3},leaf,text width=11em]
        ]
        [Enc.Dec.
            [ BART\cite{lewis20bart}{,} T5\cite{raffel2020t5}{,} Switch Transformer\cite{fedus2021switch}
            ,leaf,text width=11em]
        ]
    ]
    [App.
        [NLP
            [BERT\cite{devlin2019bert}{,}ET\cite{so19evolvedxformer}{,} Transformer-XL\cite{dai2019transformerxl}{,}Compressive Transformer\cite{rae2019compressive}{,} TENER\cite{yan2019tener}
            ,leaf,text width=22em]
        ]
        [CV
            [Image Transformer\cite{parmar2018imagexformer}{,} DETR\cite{carion20detr}{,} ViT\cite{dosovitskiy2020vit}{,} Swin Transformer\cite{liu2021swin}{,} ViViT\cite{arnab2021vivit}
            ,leaf,text width=22em]
        ]
        [Audio
            [Speech Transformer\cite{dong18speechxformer}{,} Streaming Transformer\cite{chen20streamingxformer}{,} Reformer-TTS\cite{ihm20reformertts}{,} Music Transformer\cite{huang2018music}
            ,leaf,text width=25em]
        ]
        [Multimodal
            [VisualBERT\cite{li2019visualbert}{,} VLBERT\cite{su20vlbert}{,} VideoBERT\cite{sun19videobert}{,} M6\cite{lin2021m6}{,} Chimera\cite{han2021chimera}{,} DALL-E\cite{ramesh2021dalle}{,} CogView\cite{ding2021cogview}
            ,leaf,text width=25em]
        ]
    ]
  ]
\end{forest}
\caption{Taxonomy of Transformers}
\label{taxonomy_of_xformer}
\end{figure*}

In this survey, we focus on reviewing the works on architecture modifications.
Since the attention module is the key component of Transformer, we solely describe the attention-related variants in Sec.~\ref{sec:attention} and introduce the other module-level variants in Sec.~\ref{sec:other_module}. Then Sec.~\ref{sec:beyond} describes the other architecture-level variants. Finally, we briefly review the works on pre-training in Sec.~\ref{sec:ptm} and applications in Sec.~\ref{sec:app}.
There are some comprehensive surveys on the latter two categories of work, such as pre-trained models (PTMs)~\cite{qiu2020ptms} and visual Transformers\cite{han2021surveyvisualxformer,khan2021xformersinvision}.

\section{Attention}\label{sec:attention}

Self-attention plays an important role in Transformer, but there are two challenges in practical applications.

\begin{enumerate}
  \item \textit{Complexity}. As discussion in Sec.~\ref{sec:model-analysis}, the complexity of self-attention is $\mathcal O(T^2\cdot D)$. Therefore, the attention module becomes a bottleneck when dealing with long sequences.
  \item \textit{Structural prior}. Self-attention does no assume any structural bias over inputs. Even the order information is also needed to be learned from training data. Therefore, Transformer (w/o pre-training) is usually easy to overfit on small or moderate-size data.
\end{enumerate}

The improvements on attention mechanism can be divided into several directions:
\begin{enumerate}
  \item \textit{Sparse Attention}. This line of work introduces sparsity bias into the attention mechanism, leading to reduced complexity.
  \item \textit{Linearized Attention}. This line of work disentangles the attention matrix with kernel feature maps. The attention is then computed in reversed order to achieve linear complexity.
  \item \textit{Prototype and Memory Compression}. This class of methods reduces the number of queries or key-value memory pairs to reduce the size of the attention matrix.
  \item \textit{Low-rank Self-Attention}. This line of work capture the low-rank property of self-attention.
  \item \textit{Attention with Prior}. The line of research explores supplementing or substituting standard attention with prior attention distributions.
  \item \textit{Improved Multi-Head Mechanism}. The line of studies explores different alternative multi-head mechanisms.
\end{enumerate}
We will describe these attention variants at length in the rest of this section.

\subsection{Sparse Attention}\label{sec:sparseattn}

In the standard self-attention mechanism, every token needs to attend to all other tokens.
However, it is observed that for the trained Transformers the learned attention matrix $\bA$ is often very sparse across most data points~\cite{child2019generating}.
Therefore, it is possible to reduce computation complexity by incorporating structural bias to limit the number of query-key pairs that each query attends to. Under this limitation, we just compute the similarity score of the query-key pairs according to pre-defined patterns
\begin{equation}\label{Sparse attention:position-based}
    \mathrm{\hat{\bA}}_{ij} = \begin{cases} \bq_i\bk_{j}^\top  & \text{if token }i\text{ attends to token }j,\\
    -\infty & \text{if token }i\text{ does not attend to token }j,
    \end{cases}
\end{equation}\label{eq:sparseattn}
where $\mathrm{\hat{\bA}}$ is un-normalized attention matrix. In implementation the $-\infty$ item is usually not stored in memory so as to decrease memory footprint.

From another perspective, the standard attention can be regarded as a complete bipartite graph where each query receives information from all memory nodes and updates its representation.
The sparse attention can be considered as a sparse graph where some of the connections between nodes are removed.

Based on the metrics of determining the sparse connection, we categorize these approaches into two classes: \textit{position-based} and \textit{content-based} sparse attention.

\subsubsection{Position-based Sparse Attention}\label{sec:pos_based}

In position-based sparse attention, the attention matrix is limited according to some pre-defined patterns.
Although these sparse patterns vary in different forms, we find that some of them can be decomposed into some atomic sparse patterns.

We first identify some atomic sparse patterns and then describe how these patterns are composed in some existing work. Finally, we introduce some extended sparse patterns for specific data types.

\begin{figure}[htbp]
\begin{center}
\subfigure[global\label{fig:global}]{
\includegraphics[width=0.18\linewidth]{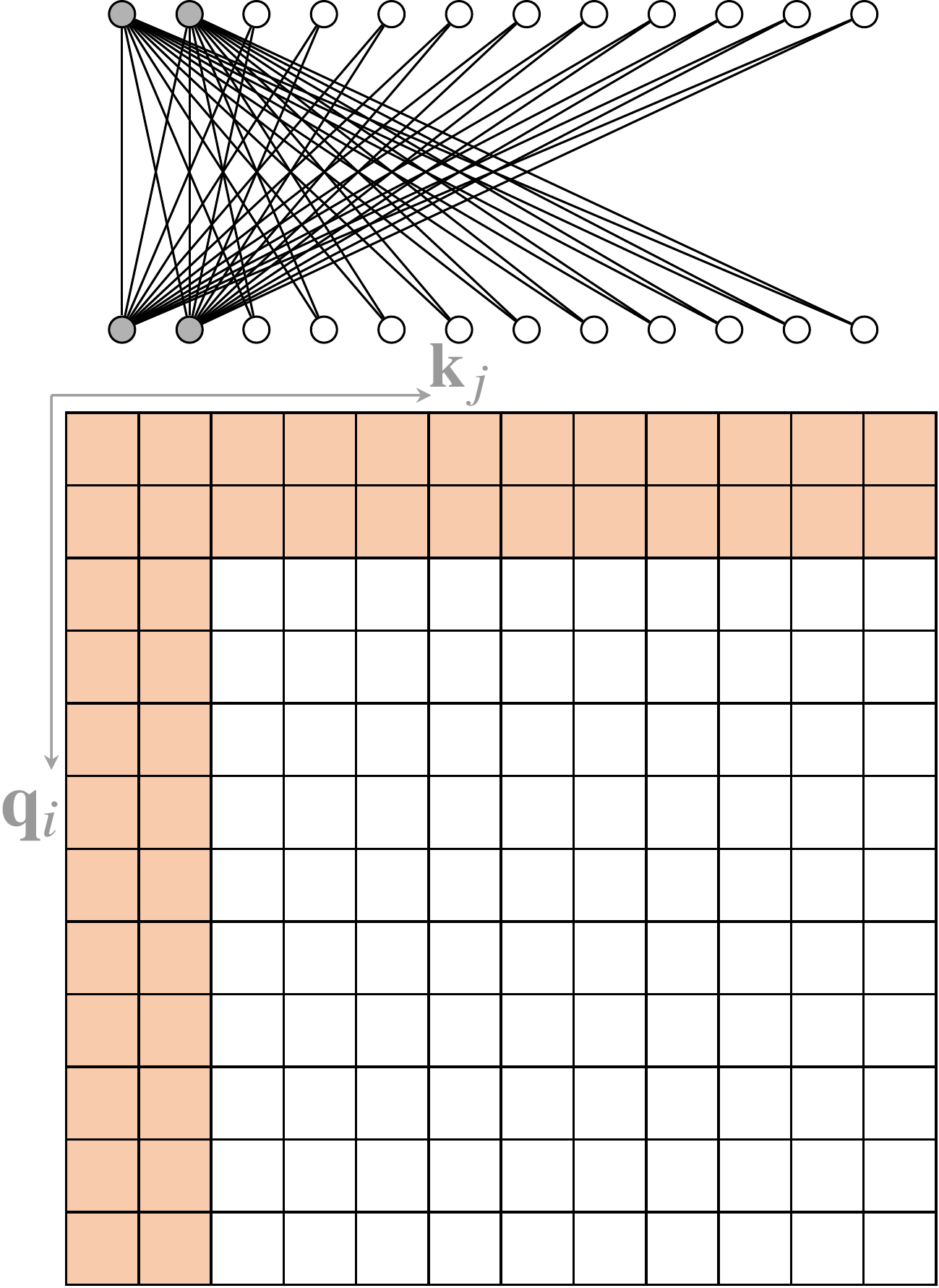}
}
\subfigure[band\label{fig:band}]{
\includegraphics[width=0.18\linewidth]{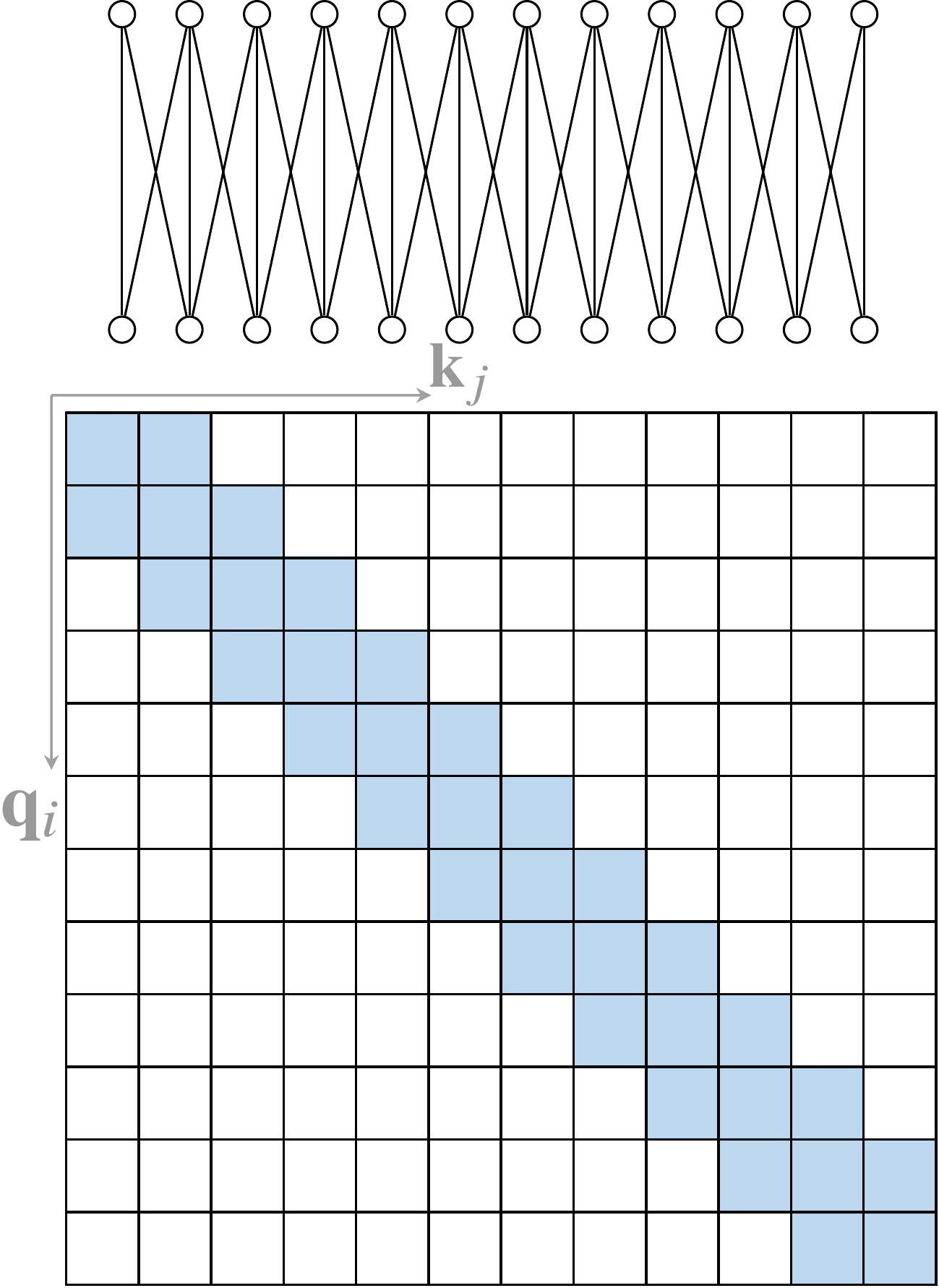}
}
\subfigure[dilated\label{fig:dilated}]{
\includegraphics[width=0.18\linewidth]{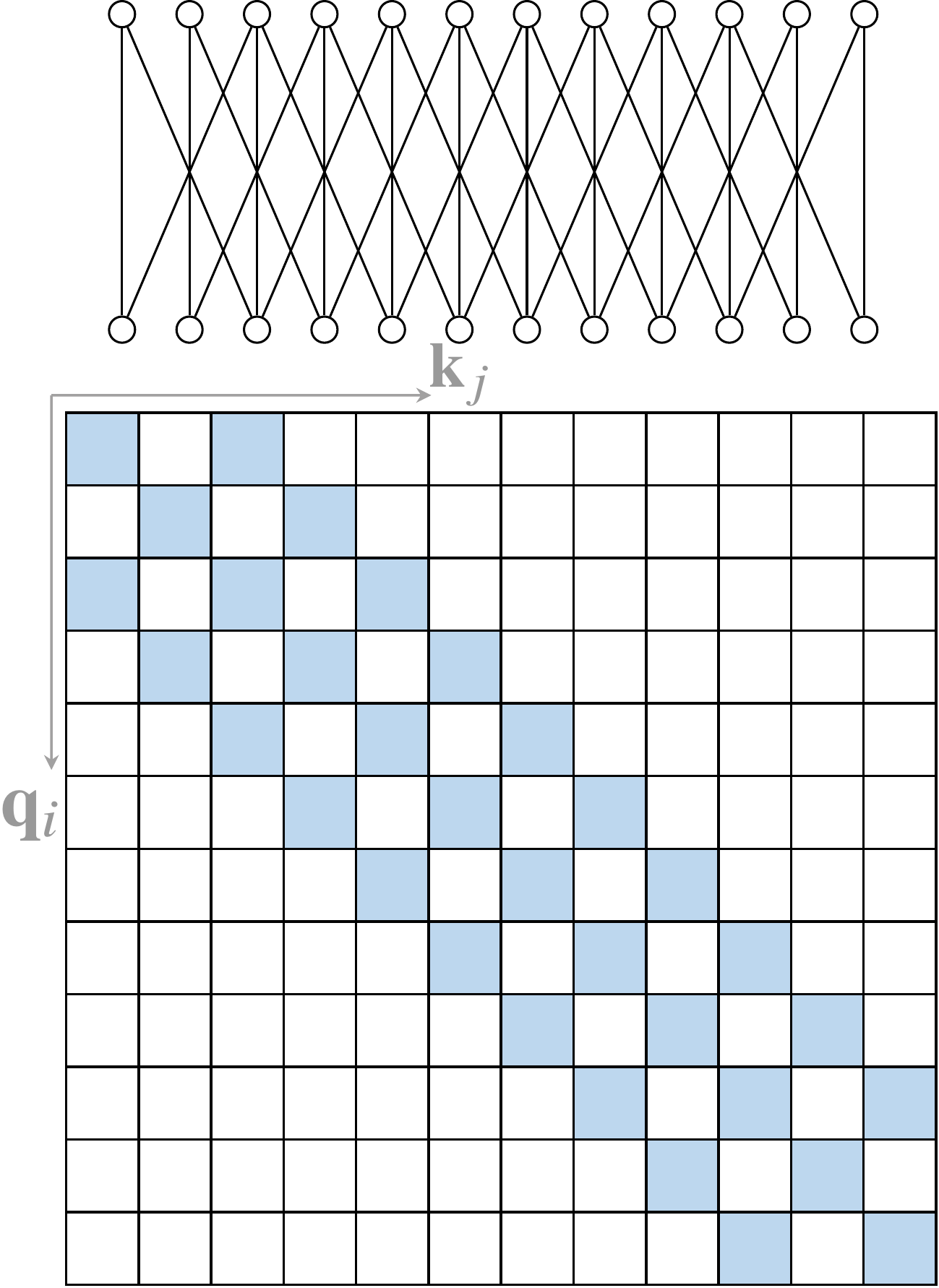}
}
\subfigure[random\label{fig:random}]{
\includegraphics[width=0.18\linewidth]{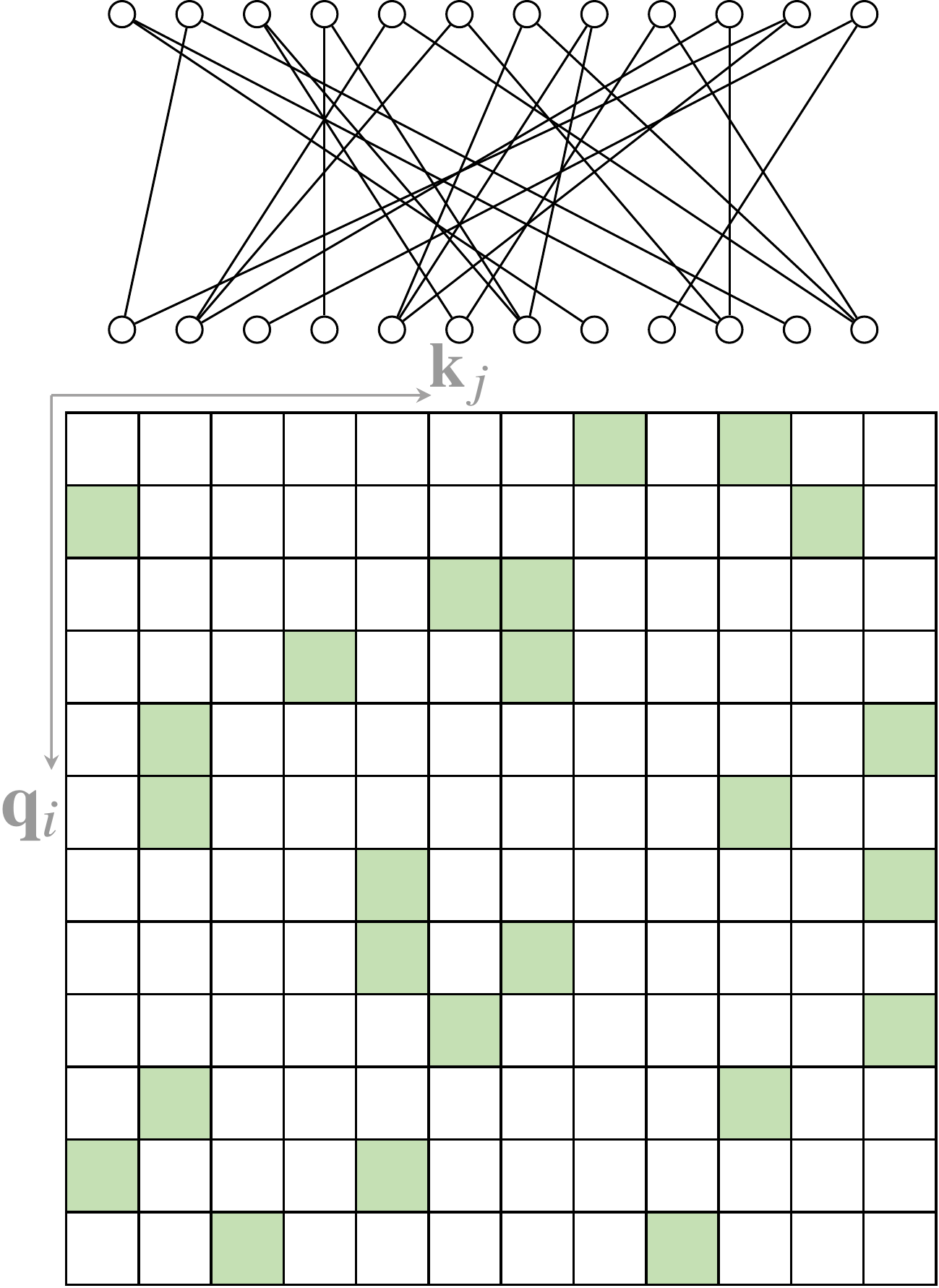}
}
\subfigure[block local\label{fig:block_local}]{
\includegraphics[width=0.18\linewidth]{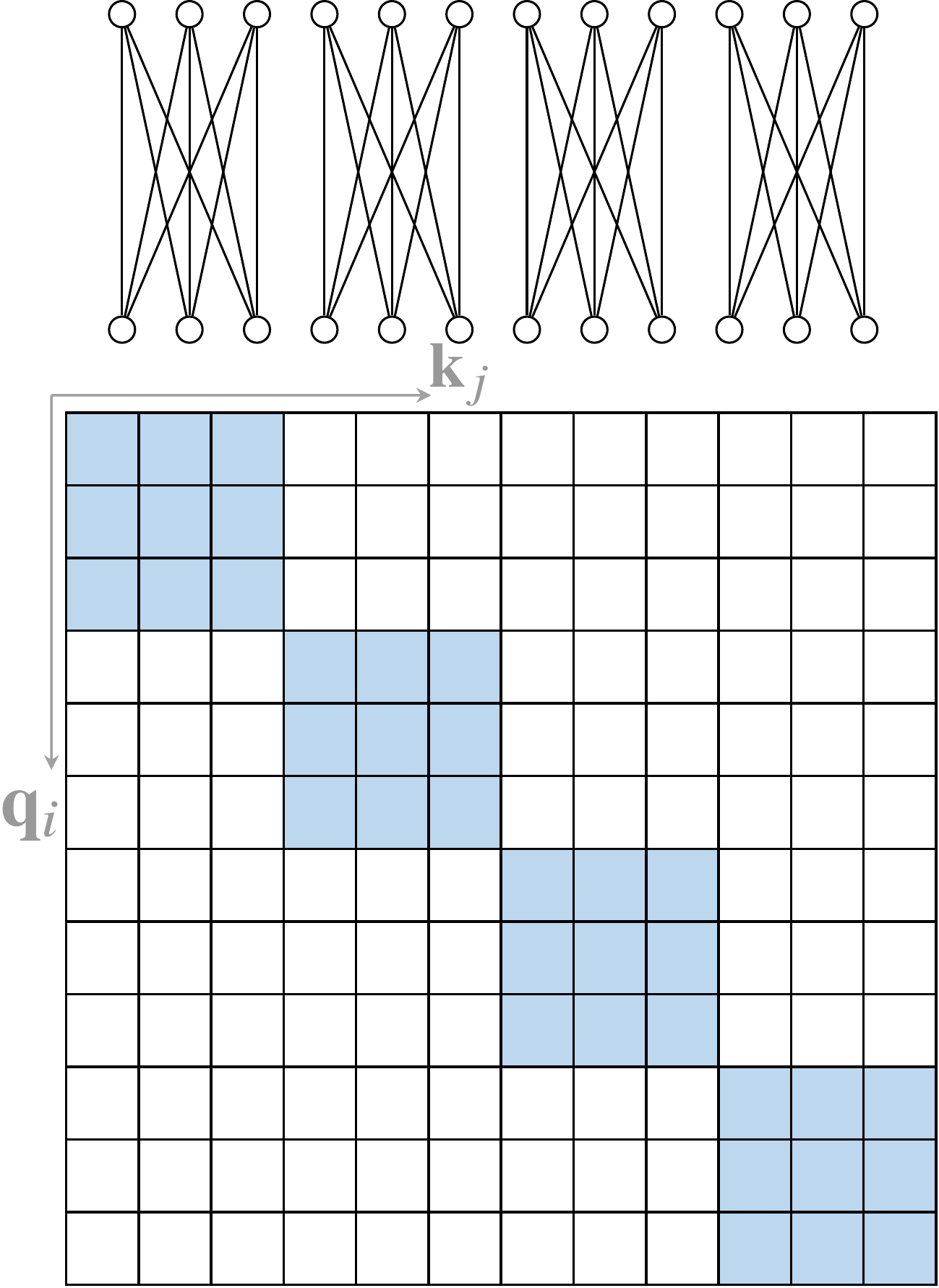}
}
\end{center}
\caption{Some representative atomic sparse attention patterns. The colored squares means corresponding attention scores are calculated and a blank square means the attention score is discarded.}\label{fig:atomic_sparse_attentions}
\end{figure}

\paragraph{4.1.1.1 Atomic Sparse Attention}
There are mainly five types of atomic sparse attention patterns, as shown in Fig. \ref{fig:atomic_sparse_attentions}.

\begin{enumerate}
  \item \textit{Global Attention}. To alleviate the degradation of the ability to model the long-range dependencies in sparse attention, one can add some global nodes\footnote{In practice, these global nodes can be selected from the sequence (internal global nodes) or virtual nodes with trainable parameters (external global nodes).} as the hub for information propagation between nodes. These global nodes can attend all nodes in the sequence and the whole sequence attend to these global nodes, as illustrated in Fig. \ref{fig:global}.
  \item \textit{Band Attention}(a.k.a \textit{sliding window attention} or \textit{local attention}). Since most data come with a strong property of locality, it is natural to restrict each query to attend to its neighbor nodes. A widely adopted class of such sparse pattern is band attention, in which the attention matrix is a band matrix as illustrated in Fig. \ref{fig:band}.
  \item \textit{Dilated Attention}. Analogous to dilated CNNs~\cite{oord16dilatedcnn}, one can potentially increase the receptive field of the band attention without increasing computation complexity by using a dilated window with gaps of dilation $w_d\ge 1$, as depicted in Fig. \ref{fig:dilated}. This can be easily extended to \textit{strided attention}, where the window size is not limited but the dilation $w_d$ is set to a large value.
  \item \textit{Random Attention}. To increase the ability of non-local interactions, a few edges are randomly sampled for each query, as illustrated in Fig. \ref{fig:random}. This is based on the observation that random graphs (e.g., Erd\H os–R\'enyi random graph) can have similar spectral properties with complete graphs that leads to a fast mixing time for random walking on graphs.
  \item \textit{Block Local Attention}. This class of attention segments input sequence into several non-overlapping query blocks, each of which is associated with a local memory block. All the queries in a query block attend to only the keys in the corresponding memory block. Fig. \ref{fig:block_local} depicts a commonly used case where the memory blocks are identical to their corresponding query blocks.
\end{enumerate}

\begin{figure}[htbp]
\begin{center}
\subfigure[Star-Transformer\label{fig:starxformer}]{
\includegraphics[width=0.18\linewidth]{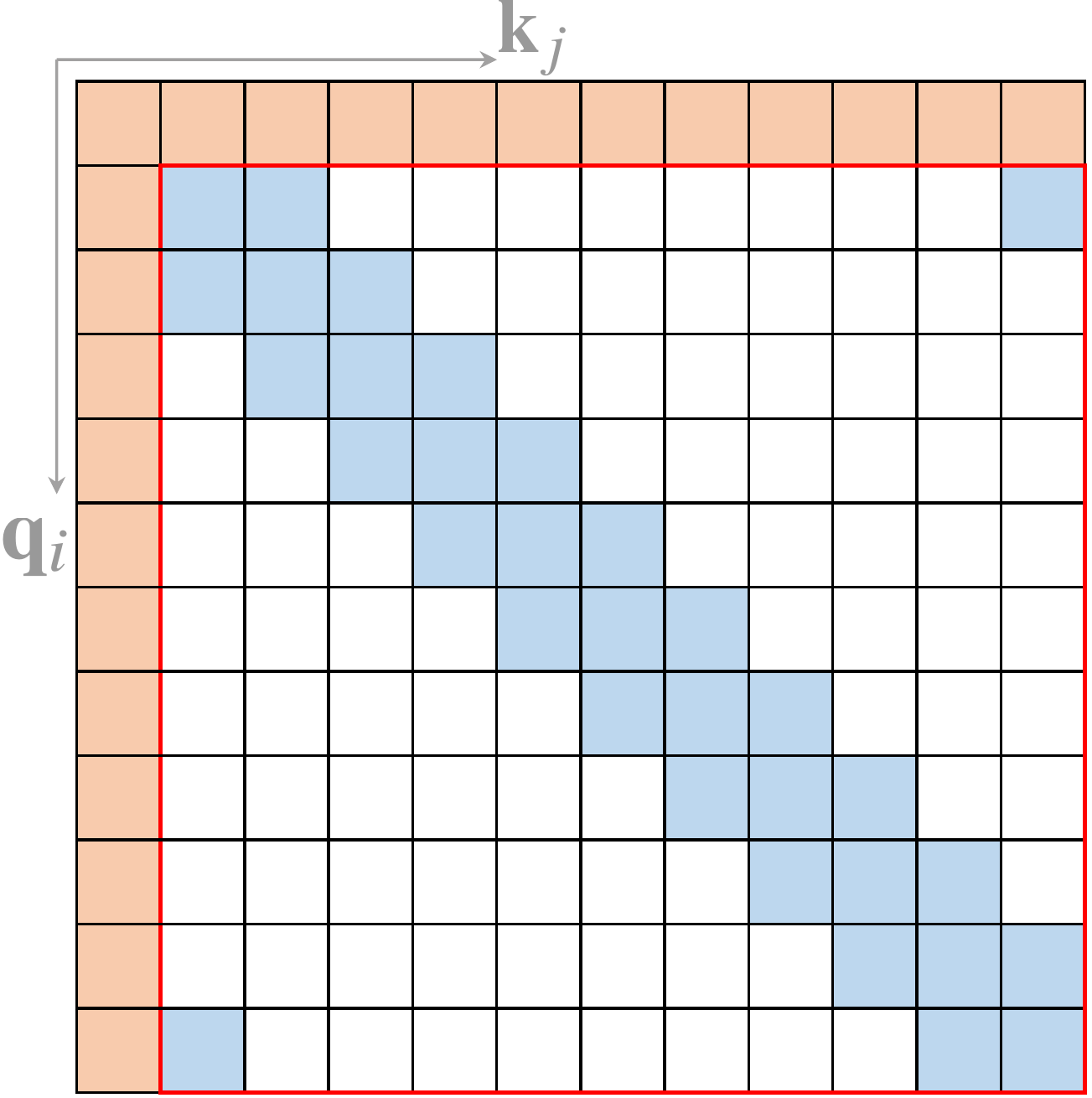}
}\quad
\quad
\subfigure[Longformer\label{fig:longformer}]{
\includegraphics[width=0.18\linewidth]{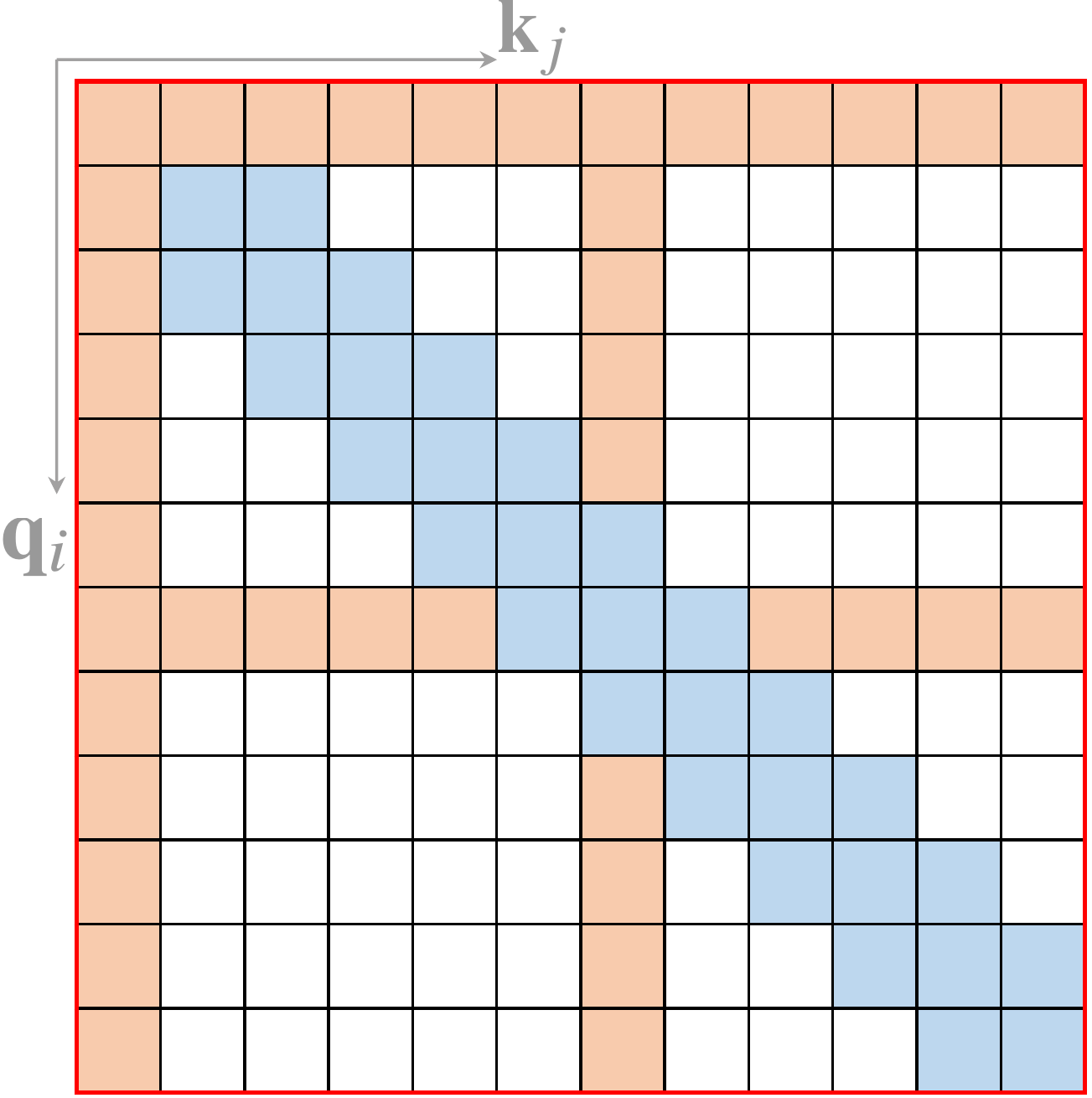}
}\quad
\subfigure[ETC\label{fig:etc}]{
\includegraphics[width=0.18\linewidth]{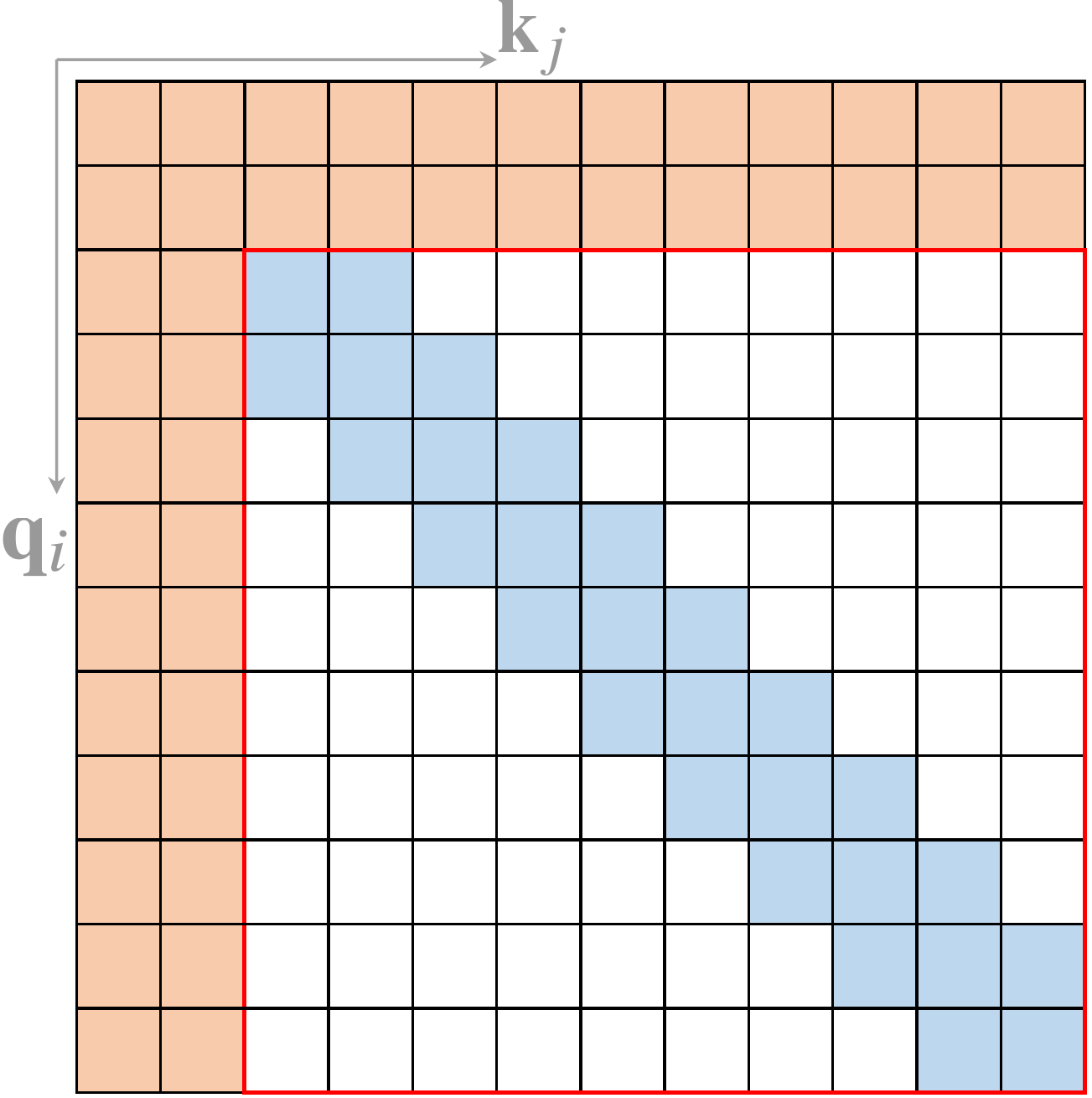}
}\quad
\subfigure[BigBird\label{fig:bigbird}]{
\includegraphics[width=0.18\linewidth]{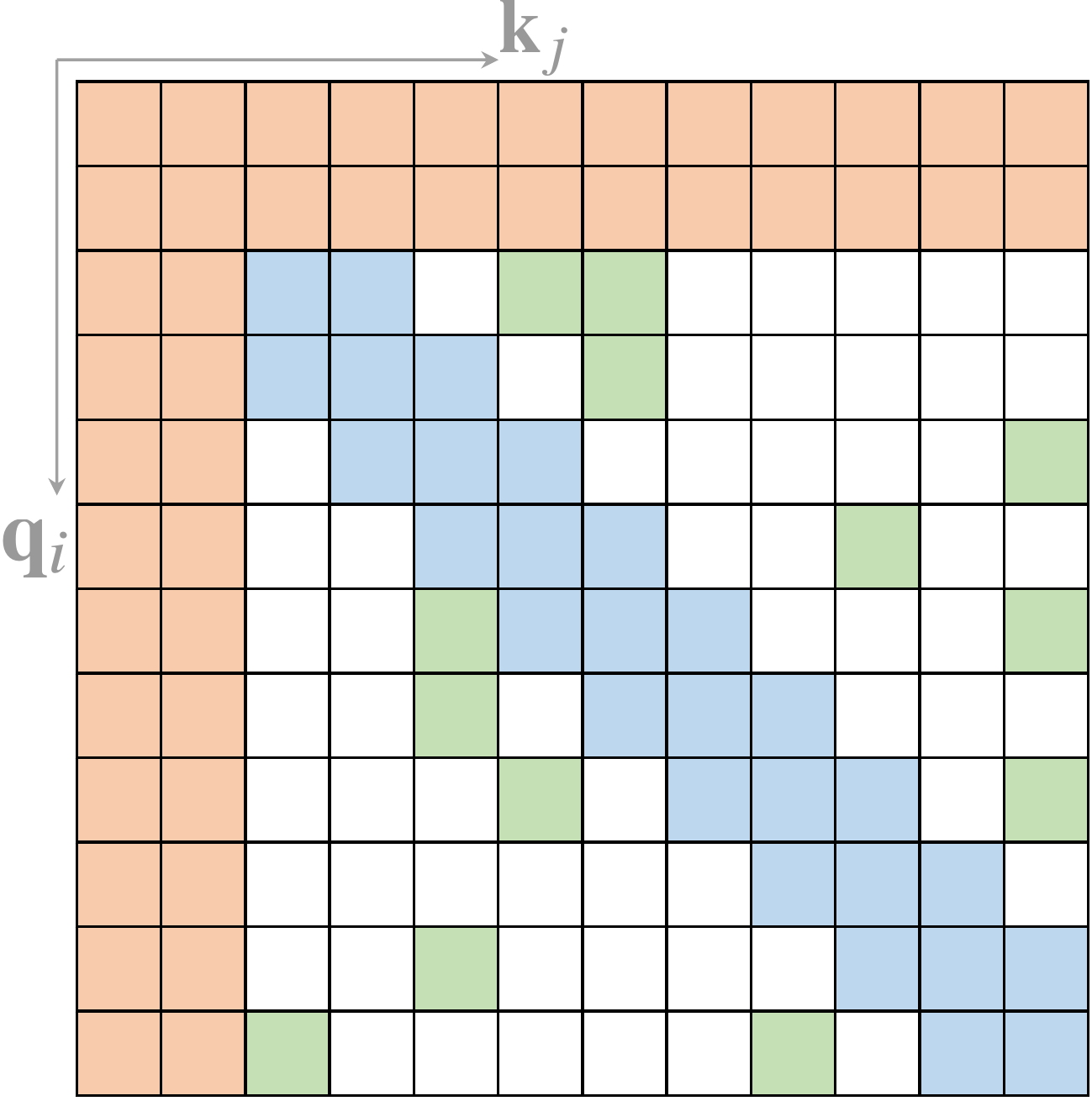}
}\quad
\caption{Some representative compound sparse attention patterns. The red boxes indicate sequence boundaries.}\label{fig:sparse_attn}
\end{center}
\end{figure}

\paragraph{4.1.1.2  Compound Sparse Attention}
Existing sparse attentions are often composed of more than one of the above atomic patterns. Fig.~\ref{fig:sparse_attn} illustrates some representative compound sparse attention patterns.

Star-Transformer~\cite{guo2019startransformer} uses a combination of band attention and global attention. Specifically, Star-Transformer just includes only a global node and a band attention with the width of 3, in which any pair of non-adjacent nodes are connected through a shared global node and adjacent nodes are connected directly with each other. This kind of sparse pattern forms a star-shaped graph among nodes.
Longformer~\cite{beltagy2020longformer} uses a combination of band attention and internal global-node attention. The global nodes are chosen to be \texttt{[CLS]} token for classification and all question tokens for Question Answering tasks. They also replace some of the band attention heads in upper layers with dilated window attention to increase the receptive field without increasing computation.
As a concurrent work to Longformer~\cite{beltagy2020longformer}, Extended Transformer Construction (ETC)~\cite{ainslie2020etc} utilizes combination of band attention and external global-node attention. ETC also includes a masking mechanism to handle structured inputs and adapt Contrastive Predictive Coding (CPC)~\cite{oord18cpc} for pre-training.
In addition to the band and global attention, BigBird~\cite{zaheer2020big} uses additional random attention to approximate full attention. Their theoretical analysis also reveals that the usage of a sparse encoder and sparse decoder can simulate any Turing Machine, which explains the success of those sparse attention models.

Sparse Transformer~\cite{child2019generating} uses a factorized attention where different sparse patterns are designed for different types of data. For data with a periodic structure (e.g., images), it uses a composition of band attention and strided attention. Whereas for data without a periodic structure (e.g., text), it uses a composition of block local attention combined with global attention, where global nodes are from fixed positions in the input sequence.

\paragraph{4.1.1.3  Extended Sparse Attention}
Apart from the above patterns, some existing studies have explored extended sparse patterns for specific data types.

\begin{figure}[htbp]
\begin{center}
\subfigure[BPT\label{fig:bpt}]{
\includegraphics[width=0.40\linewidth]{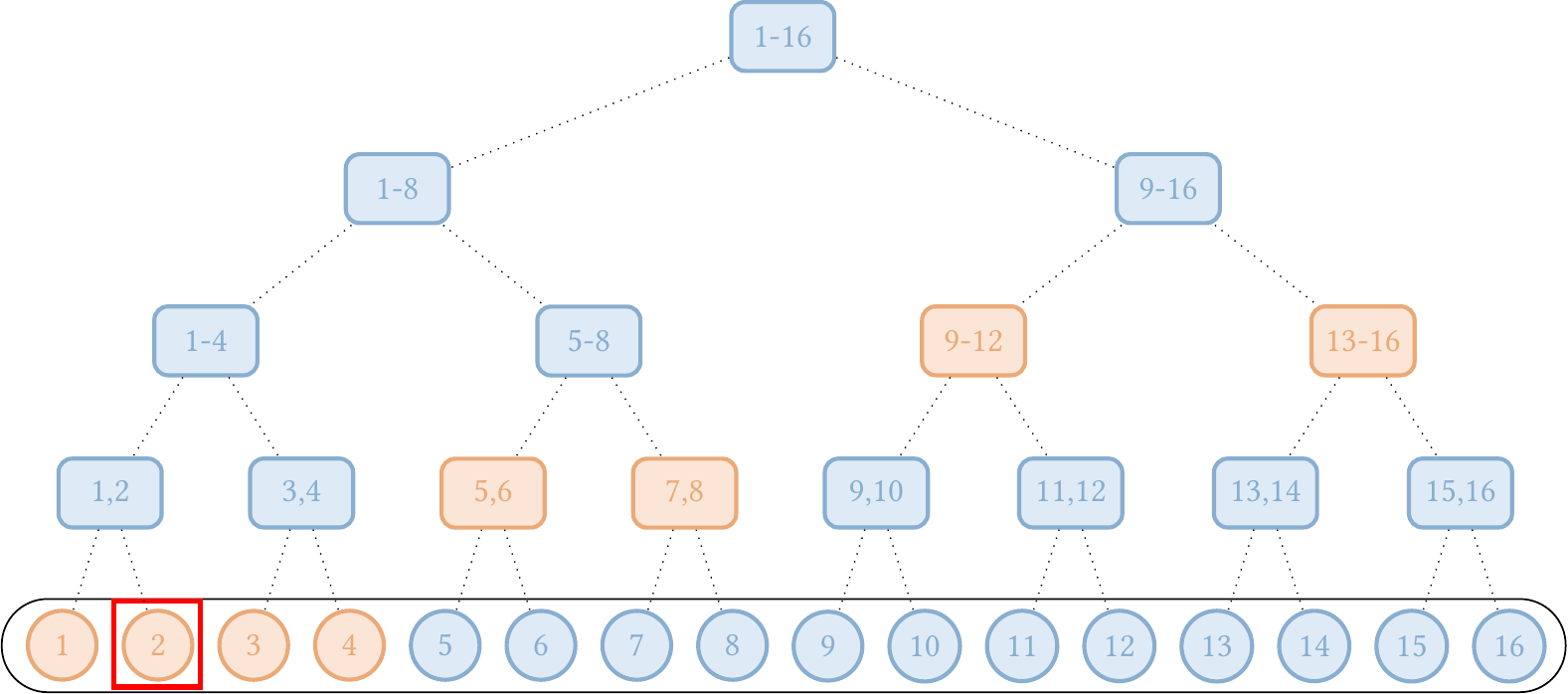}
}\quad
\subfigure[block local (2D)\label{fig:block_local_2d}]{
\includegraphics[width=0.18\linewidth]{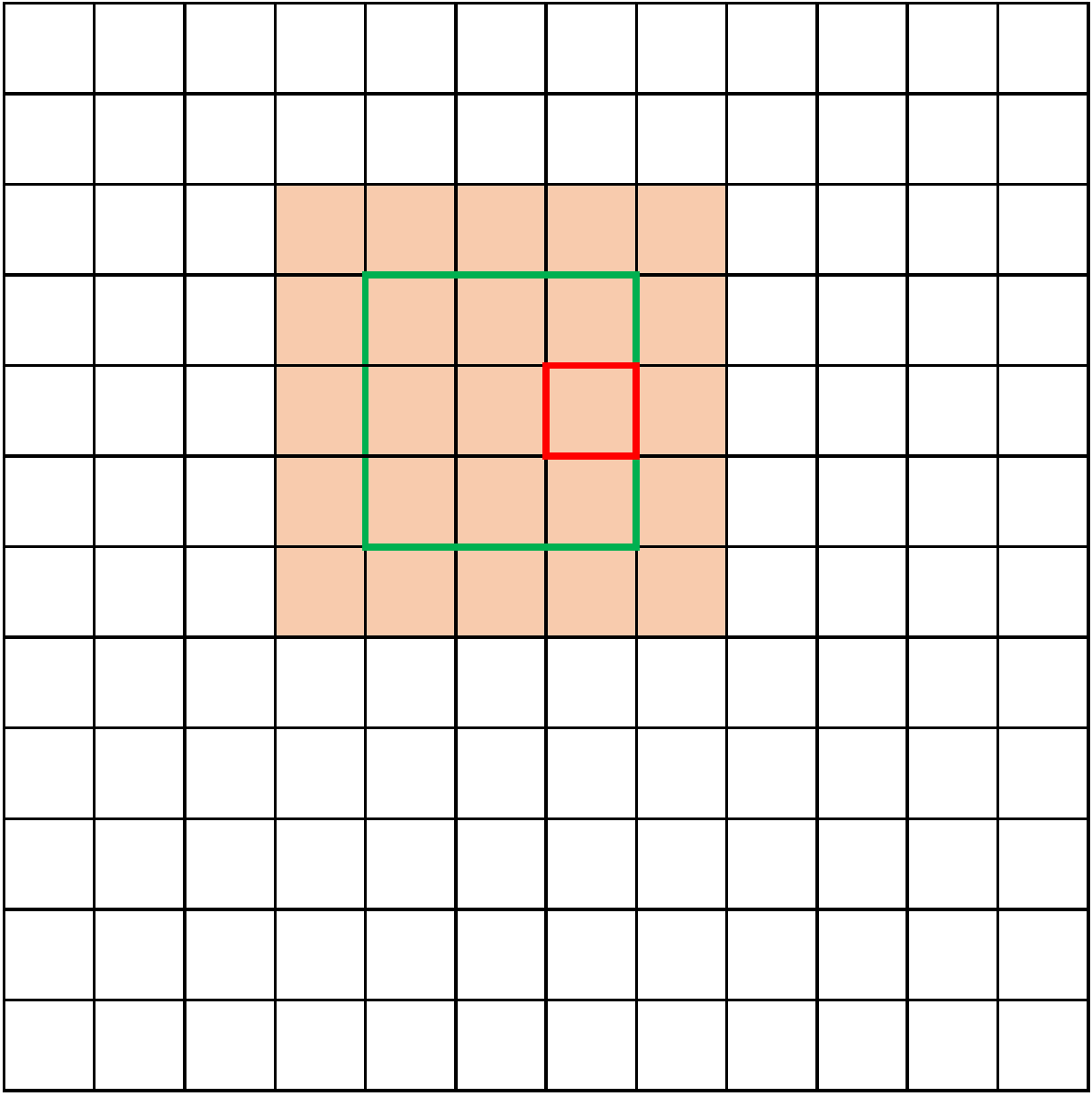}
}\quad
\subfigure[axial (2D)\label{fig:axial}]{
\includegraphics[width=0.18\linewidth]{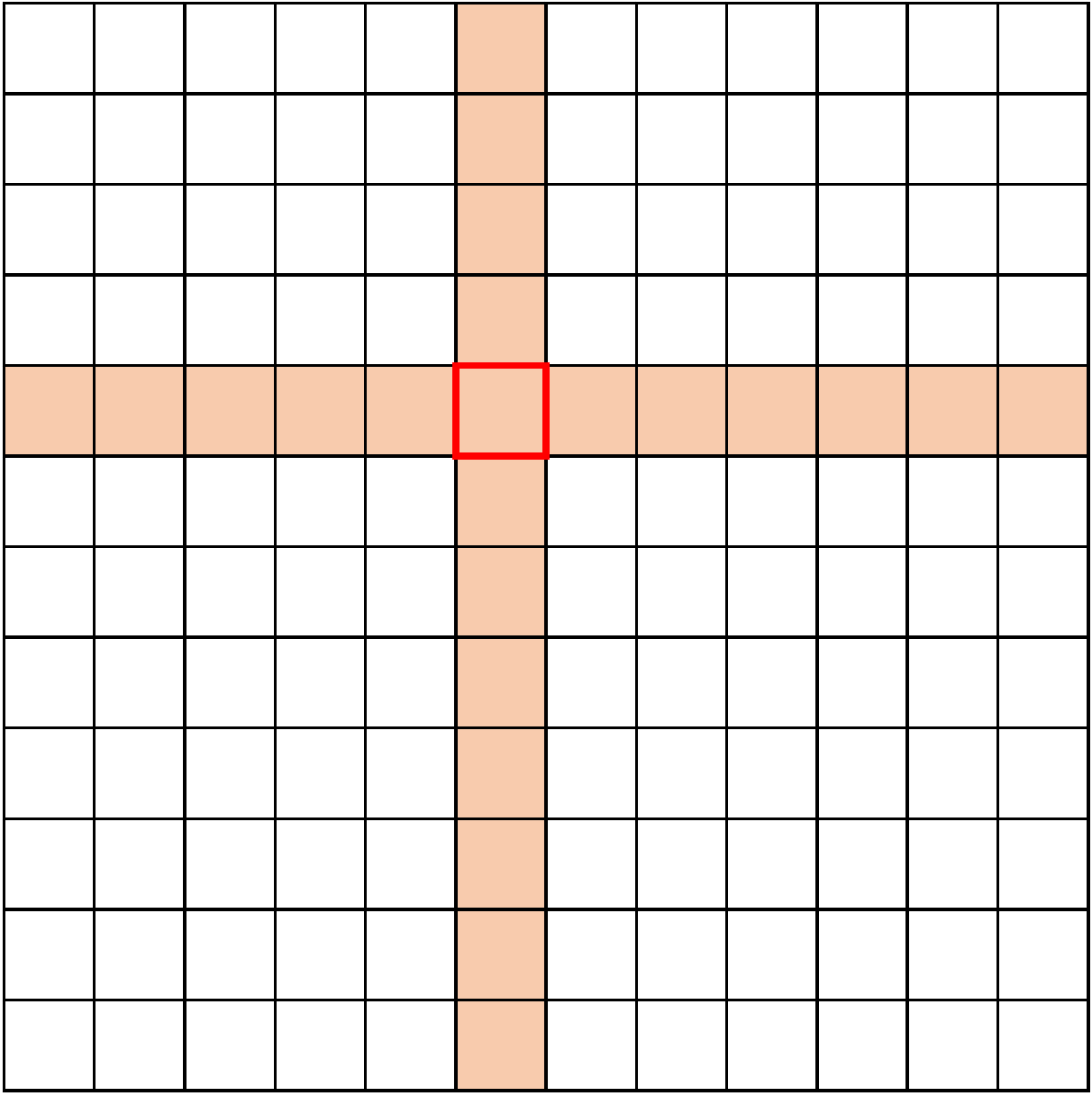}
}
\caption{Other types of sparse attentions. The red box indicates the query position, and the orange nodes/squares means corresponding tokens are attended to by the query.}\label{fig:other_sparse_attn}
\end{center}
\end{figure}

For text data, BP-Transformer~\cite{ye2019bptransformer} constructs a binary tree where all tokens are leaf nodes and the internal nodes are span nodes containing many tokens. The edges in this graph are constructed so that each leaf node is connected to its neighbor leaf nodes and higher-level span nodes containing tokens from a longer distance. This approach can be seen as an extension of global attention, where global nodes are hierarchically organized and any pair of tokens are connected with paths in the binary tree. An abstract view of this method is illustrated in Fig. \ref{fig:bpt}.

There are also some extensions for vision data. Image Transformer~\cite{parmar2018imagexformer} explores two types of attention: (1) flattening image pixels in raster-scan order and then applying block local sparse attention. (2) 2D block local attention, where query blocks and memory blocks are arranged directly in 2D plate, as depicted in Fig. \ref{fig:block_local_2d}. As another example of sparse pattern on vision data, Axial Transformer~\cite{ho19axialxformer} applies independent attention modules over each axis of the image. Each attention module mixes information along one axis while keeping information along the other axis independent, as illustrated in Fig. \ref{fig:axial}. This can be understood as horizontally and vertically flattening image pixels in raster-scan order and then applying strided attention with gaps of image width and height, respectively.

\subsubsection{Content-based Sparse Attention}
Another line of work creates a sparse graph based on input content, i.e., the sparse connections are conditioned on inputs.

A straightforward way of constructing a content-based sparse graph is to select those keys that are likely to have large similarity scores with the given query. To efficiently construct the sparse graph, we can recur to Maximum Inner Product Search (MIPS) problem, where one tries to find the keys with maximum dot product with a query without computing all dot product terms. Routing Transformer~\cite{roy2020efficient} uses k-means clustering to cluster both queries $\{\bq_i\}_{i=1}^T$ and keys $\{\bk_i\}_{i=}^T$  on the same set of centroid vectors $\{\mathbf\mu_i\}_{i=1}^k$. Each query only attends to the keys that belong to the same cluster.
During training, the cluster centroid vectors are updated using the exponentially moving average of vectors assigned to it, divided by the exponentially moving average of cluster counts:
\begin{align}
\tilde\mu&\leftarrow \lambda\tilde\mu+(1-\lambda)\left(\sum_{i:\mu(\bq_i)=\mu}\bq_i+\sum_{j:\mu(\bk_j)=\mu}\bk_j\right),\\
c_\mu&\leftarrow \lambda c_\mu+(1-\lambda)|\mu|,\\
\mu&\leftarrow \frac{\tilde\mu}{c_\mu},
\end{align}
where $|\mu|$ denotes the number of vectors currently in cluster $\mu$ and $\lambda\in(0,1)$ is a hyperparameter.

Let $\mathcal{P}_i$ denote the set of indices of keys that the $i$-th query attend to. $\mathcal{P}_i$ in Routing Transformer is defined as
\begin{equation}
    \mathcal{P}_i=\{j: \mu(\bq_i)=\mu(\bk_j) \}.
\end{equation}

Reformer~\cite{kitaev2020reformer} uses locality-sensitive hashing (LSH) to select key-value pairs for each query. The proposed LSH attention allows each token to attend only to the tokens within the same hashing bucket. The basic idea is to use an LSH function to hash queries and keys into several buckets, with similar items fall in the same bucket with high probability. Specifically, they use the random matrix method for the LSH function. Let $b$ be the number of buckets, given a random matrix $R$ of size $[D_k,b/2]$, the LSH function is computed by :
\begin{equation}
    h(x)=\argmax([xR;-xR]).
\end{equation}

The LSH attention allows the $i$-th query to attend only to key-value pairs with indices
\begin{equation}
    \mathcal{P}_i=\{j: h(\bq_i)=h(\bk_j) \}.
\end{equation}

Sparse Adaptive Connection (SAC)~\cite{li2020sac} views the input sequence as a graph and learns to construct attention edges to improve task-specific performances using an adaptive sparse connection. SAC uses an LSTM edge predictor to construct edges between tokens. With no ground truth for edges, the edge predictor is trained with reinforcement learning.

Sparse Sinkhorn Attention~\cite{tay2020sparse} first splits queries and keys into several blocks and assigns a key block to each query block. Each query is only allowed to attend to the keys in the key block that is assigned to its corresponding query block. The assignment of key blocks is controlled by a sorting network, which uses Sinkhorn normalization to produce a doubly stochastic matrix as the permutation matrix representing the assignment. They use this content-based block sparse attention along with block local attention introduced in Sec.~\ref{sec:pos_based} to enhance the ability of the model to model locality.

\subsection{Linearized Attention}\label{sec:kernel}

\begin{figure}[htbp]
\begin{center}
\subfigure[standard self-attention]{\label{fig:standard_complexity}
\includegraphics[width=0.45\textwidth]{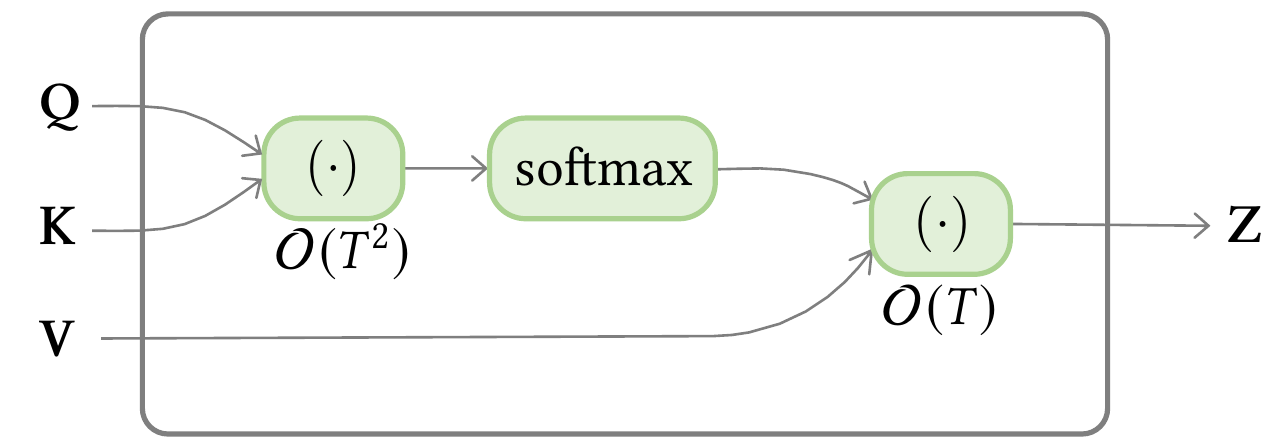}
}\hspace{.2in}
\subfigure[linearized self-attention]{\label{fig:linearized_complexity}
\includegraphics[width=0.45\textwidth]{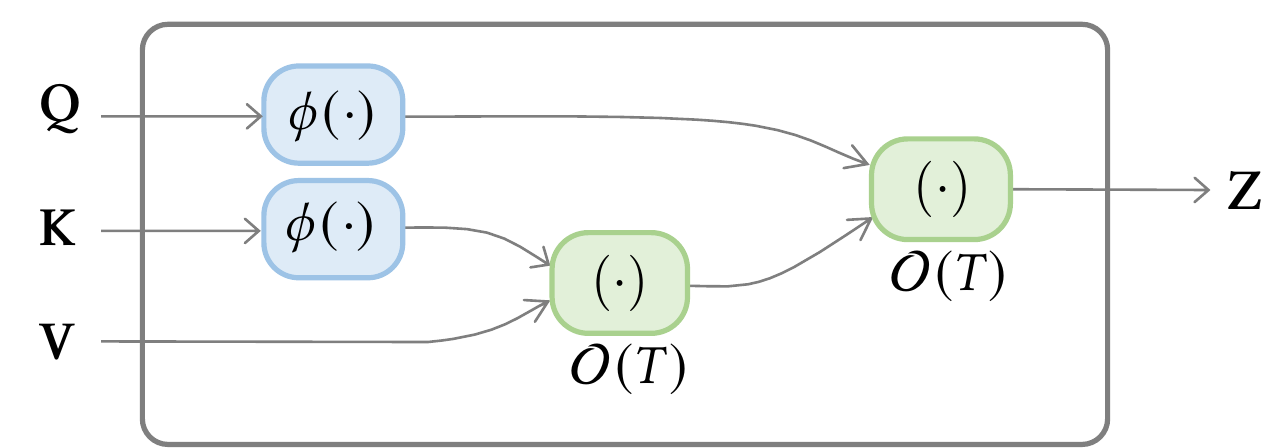}
}
\caption{Illustration of complexity difference between standard self-attention and linearized self-attention.}
\end{center}
\end{figure}

Assuming $\bQ,\bK,\bV\in \mathbb{R}^{T\times D}$, the complexity of computing $\softmax(\bQ\bK^\top)\bV$ is quadratic w.r.t. sequence length $T$, as illustrated in Fig. \ref{fig:standard_complexity}. If $\softmax(\bQ\bK^\top)$ can be disentangled into $\bQ'\bK'^\top$, we can compute $\bQ'\bK'^\top\bV$ in reversed order (i.e., $\bQ'\left(\textcolor[rgb]{0,0,1}{\bK'^\top\bV}\right)$), leading to a complexity of $\mathcal{O}(T)$.

Let $\hat{\bA} = \exp(\bQ\bK^\top)$ denote un-normalized attention matrix, and $\exp(\cdot)$ is applied element-wise, the regular attention can be rewritten as $\bZ={\bD}^{-1}\hat{\bA}\bV$, where $\bD = \mathrm{diag}(\hat{\bA}\mathbf{1}_T^\top)$; $\mathbf{1}_T^\top$ is the all-ones column vector of length $T$; $\mathrm{diag}(\cdot)$ is a diagonal
matrix with the input vector as the diagonal.

Linearized attention is a class of methods that approximate or replace the unnormalized attention matrix $\exp(\bQ\bK^\top)$ with $\phi(\bQ)\phi(\bK)^\top$, where $\phi$ is a feature map that is applied in row-wise manner. Hence the computation of unnormalized attention matrix can be linearized by computing $\phi(\bQ)\left(\textcolor[rgb]{0,0,1}{\phi(\bK)^\top\bV}\right)$\footnote{Similarly, the partition term $\bD$ can be computed with $\phi(\bQ)\left(\textcolor[rgb]{0,0,1}{\phi(\bK)^\top\mathbf{1}_T^\top}\right)$ in linear time.}, as illustrated in Fig. \ref{fig:linearized_complexity}.

To gain further insights into linearized attention, we derive the formulation in vector form. We consider a general form of attention
\begin{equation}
    \bz_i=\sum_j \frac{\mathrm{sim}(\bq_i, \bk_j)}{\sum_{j'}\mathrm{sim}(\bq_i, \bk_{j'})}\bv_j,
\end{equation}
where $\mathrm{sim}(\cdot,\cdot)$ is a scoring function measuring similarity between input vectors. In vanilla Transformer, the scoring function is the exponential of inner product $\exp(\langle\cdot,\cdot\rangle)$. A natural choice of $\mathrm{sim}(\cdot,\cdot)$ is a kernel function $\mathcal{K}(\bx,\by)=\phi(\bx)\phi(\by)^\top$, which leads to
\begin{align}
    \bz_i&=\sum_j \frac{\phi(\bq_i) \phi(\bk_j)^\top}{\sum_{j'} \phi(\bq_i) \phi(\bk_{j'})^\top}\bv_j\\
    &=\frac{\phi(\bq_i) \textcolor[rgb]{0,0,1}{\sum_j\phi(\bk_j)\otimes\bv_j}}{\phi(\bq_i)\textcolor[rgb]{0,0,1}{\sum_{j'} \phi(\bk_{j'})^\top}},\label{eq:linearized}
\end{align}
where $\otimes$ denotes outer product of vectors. Based on this formulation, attention can be linearized by first computing the highlighted terms $\sum_j\phi(\bk_j)\otimes\bv_j$ and$\sum_{j'} \phi(\bk_{j'})^\top$. This could be especially beneficial for autoregressive attention, as the cumulative sums $\bS_i=\sum_{j=1}^{i} \phi(\bk_j)\otimes\bv_j$ and $\bu_i=\sum_{j=1}^{i} \phi(\bk_j)$ can be computed from $\bS_{i-1}$ and $\bu_{i-1}$ in constant time. The effectively enables Transformer decoders to run like RNNs.

An interpretation of Eq. \eqref{eq:linearized} is that the model maintains a \textit{memory matrix} by aggregating \textit{associations} represented by outer products of (feature mapped) keys and values, and then retrieve a value by multiplying the memory matrix with feature mapped query with proper normalization. There are two key components in this approach: (1) feature map $\phi(\cdot)$, and (2) aggregation rule.

\subsubsection{Feature Maps}
Linear Transformer~\cite{katharopoulos2020linearxformer} propose to use a simple feature map $\phi_i(\bx)=\mathrm{elu}(x_i)+1$. This feature map does not aim to approximate dot product attention, but is empirically proved to perform on par with the standard Transformer.

Performer~\cite{choromanski2020masked,choromanski2020rethinking} uses random feature maps that approximate the scoring function of Transformer. The random feature maps take functions $f_1,\cdots,f_l:\mathbb R\rightarrow \mathbb R$ and $h:\mathbb{R}^D\rightarrow \mathbb R$.
\begin{equation}
    \phi(\bx) = \frac{h(\bx)}{\sqrt{m}}[f_1(\omega_1^\top \bx),\cdots,f_m(\omega_m^\top \bx),\cdots,f_l(\omega_1^\top \bx),\cdots,f_l(\omega_m^\top \bx)],
\end{equation}
where $\omega_1,\cdots,\omega_m\stackrel{\text{iid}}{\sim} \mathcal D$ are drawn from some distribution $\mathcal D\in\mathcal P(\mathbb{R}^D)$.

The first version of Performer~\cite{choromanski2020masked} is inspired from the random Fourier feature map~\cite{rahmi07random} that was originally used to approximate Gaussian kernel. It uses trigonometric functions with $h(\bx)=\exp(\frac{\|\bx\|^2}{2}), l=2, f_1=\sin, f_2=\cos$. This approach has also been used in Random Feature Attention (RFA)~\cite{peng2021random}, with the difference that $h(\bx)$ is set to $1$ as the queries and keys are $\ell_2$-normalized before applying the feature map.

Although the trigonometric random feature map leads to an unbiased approximation, it does not guarantee non-negative attention scores and thus could lead to unstable behaviors and abnormal behaviors. To mitigate this issue, the second version of Performer~\cite{choromanski2020rethinking} proposes positive random feature maps, which uses $h(\bx)=\exp(-\frac{\|\bx\|^2}{2}),l=1,f_1=\exp$ and thus guarantees unbiased and non-negative approximation of dot-product attention. This approach is more stable than \citet{choromanski2020masked} and reports better approximation results.

In addition to using random feature maps to approximate standard dot product attention, \citet{peng2021random} and \citet{choromanski2020rethinking} also explore approximating order-1 arc-cosine kernel with $h(\bx)=1,l=1,f_1=\mathrm{ReLU}$. This feature map has been show to be effective in various tasks including machine translation and protein sequence modeling.

\citet{schlag21fastweight} design a feature map that aims at facilitating orthogonality in feature space. Specifically, given an input $\bx\in\mathbb{R}^D$, the feature map $\phi:\mathbb{R}^D\rightarrow \mathbb{R}^{2\nu D}$ is defined by the partial function
\begin{equation}
    \phi_{i+2(j-1)D}(\bx)=\mathrm{ReLU}([\bx,-\bx])_i\mathrm{ReLU}([\bx,-\bx])_{i+j}\quad\text{for }i=1,\cdots,2D,j=1,\cdots,\nu.
\end{equation}

\subsubsection{Aggregation Rule}
In Eq. \eqref{eq:linearized} the associations $\{\phi(\bk)_j\otimes\bv_j\}$ are aggregated into the memory matrix by simple summation. This is adopted by several studies~\cite{katharopoulos2020linearxformer,choromanski2020masked,choromanski2020rethinking}. However, it could be more beneficial for the network to selectively drop associations as new associations are added to the memory matrix.

RFA~\cite{peng2021random} introduces a gating mechanism to the summation to model local dependency in sequence data. Specifically, when adding a new association to the memory matrix $\bS$, at a particular time step, they weigh $\bS$ by a learnable, input-dependent scalar $g$, and the new association by $(1-g)$ (and a similar mechanism to $\bu$). With this modification, history associations are exponentially decayed and recent context is favored in each timestep.

\citet{schlag21fastweight} argue that simple summation limits the capacity of the memory matrix and thus propose to enlarge the capacity in a write-and-remove fashion. Specifically, given a new input key-value pair $(\bk_i, \bv_i)$, the model first retrieve the value $\bar{\bv}_i$ currently associated with $\bk_i$ using matrix multiplication. It then writes to the memory matrix a convex combination of $\bar{\bv}_i$ and $\bv_i$, using a input-dependent gating scalar $g$, and removes the association $\bar{\bv}_i$. They also propose \textit{sum normalization} (normalizing $\phi(\bq_i),\phi(\bk_i)$ by the sum of their components before updating the memory matrix) instead of normalizing with the denominator in Eq. \eqref{eq:linearized} for this aggregation rule.

\subsection{Query Prototyping and Memory Compression}

Apart from using sparse attention or kernel-based linearized attention, one could also reduce the complexity of attention by reducing the number of queries or key-value pairs, which leads to \textit{query prototyping} and \textit{memory compression}\footnote{The key-value pairs are often referred to as a key-value memory (hence the name memory compression).} methods, respectively.

\subsubsection{Attention with Prototype Queries}
\begin{figure}[htbp]
\begin{center}
\subfigure[Query prototyping]{\label{fig:query_prototype}
\includegraphics[height=1.5in]{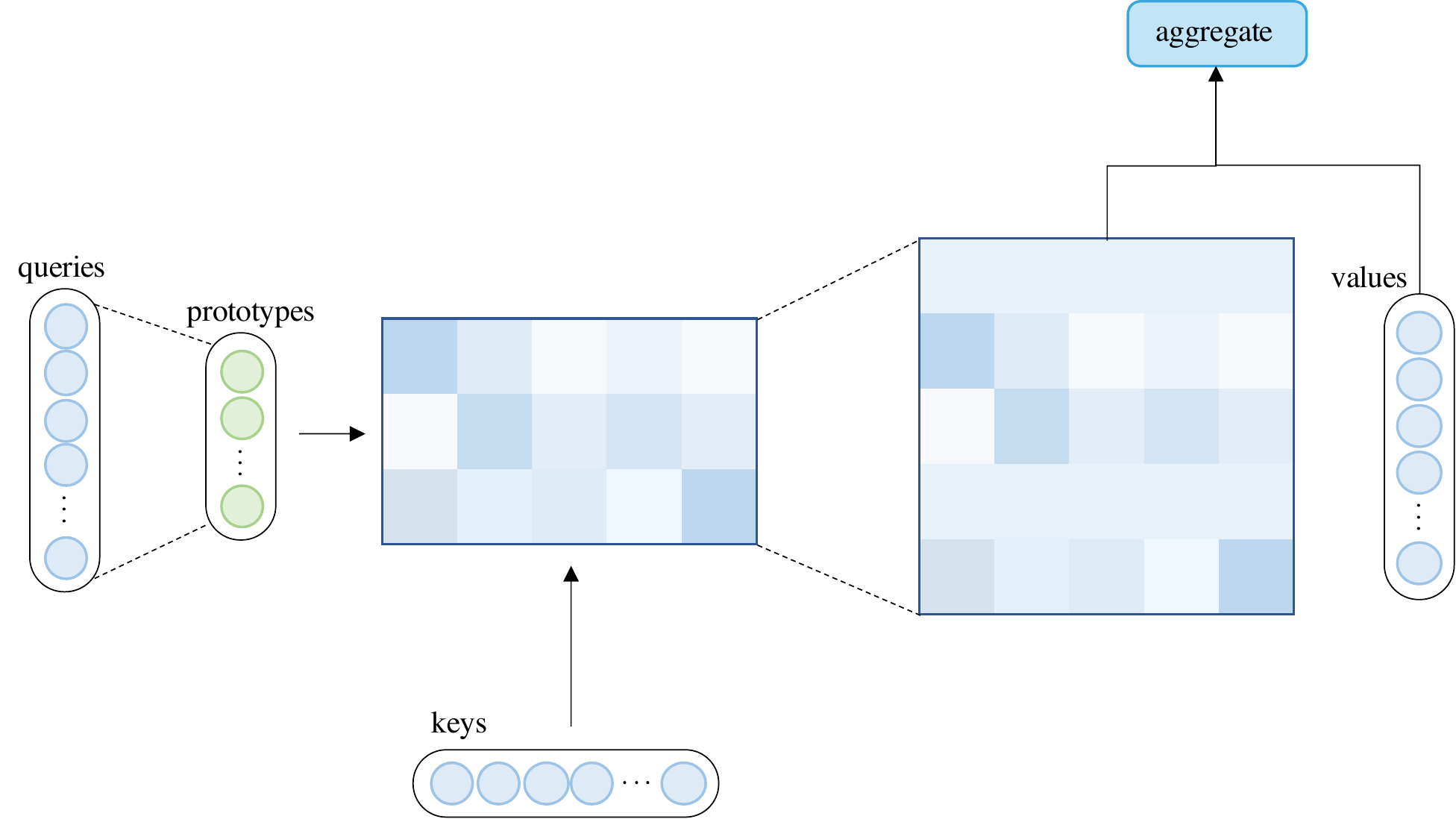}
}\hspace{.2in}
\subfigure[Memory compression]{\label{fig:compressed_mem}
\includegraphics[height=1.5in]{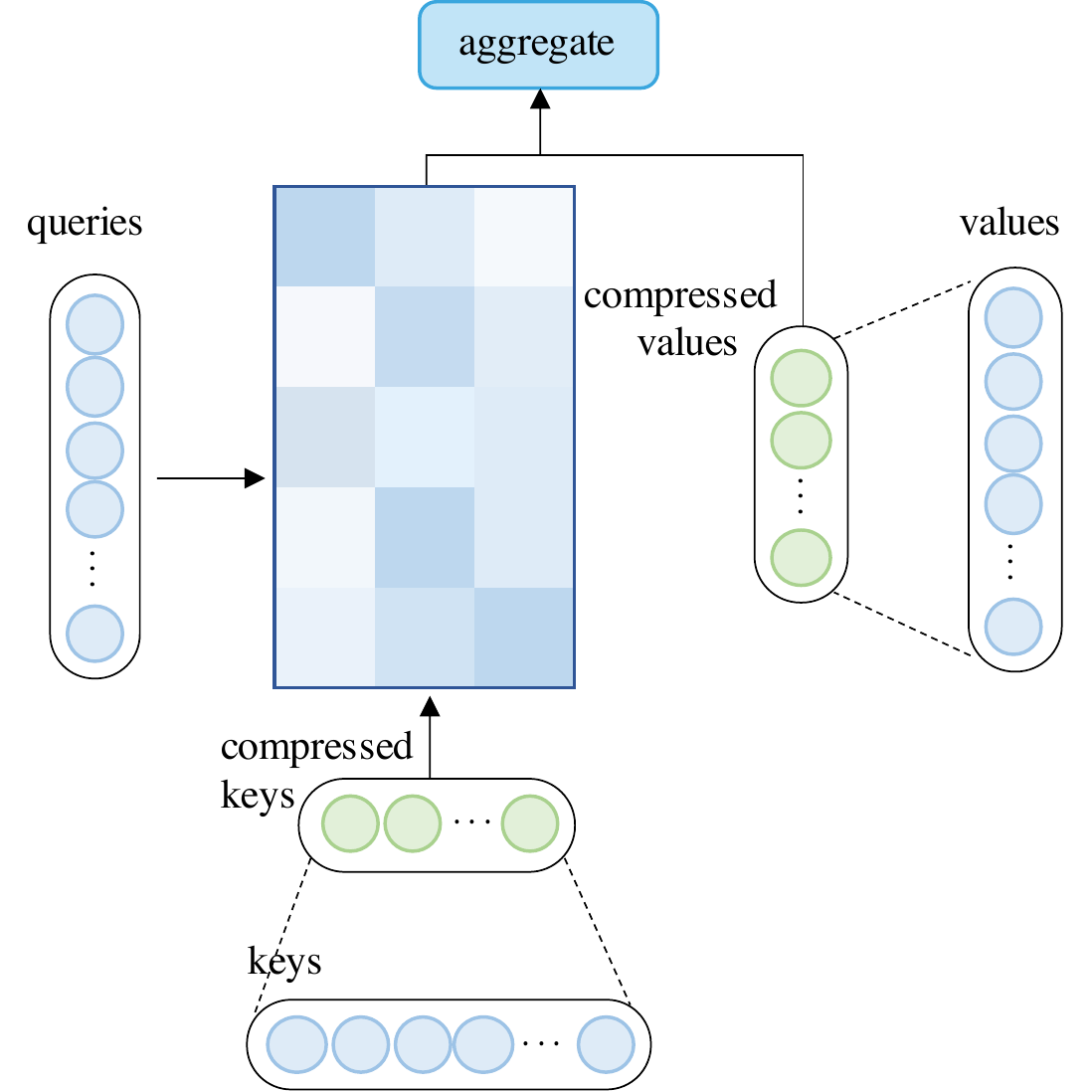}
}
\caption{Query prototyping and memory compression.}\label{fig:prototype_mem_compress}
\end{center}
\end{figure}

In query prototyping, several prototypes of queries serve as the main source to compute attention distributions. The model either copies the distributions to the positions of represented queries or filling those positions with discrete uniform distributions. Fig. \ref{fig:query_prototype} illustrates the computing flow of query prototyping.

Clustered Attention~\cite{vyas2020clusteredattn} groups queries into several clusters and then computes attention distributions for cluster centroids. All queries in a cluster share the attention distribution calculated with the corresponding centroid.

Informer~\cite{zhou20informer} selects prototypes from queries using explicit query sparsity measurement, which is derived from an approximation of the Kullback-Leibler divergence between the query's attention distribution and the discrete uniform distribution. Attention distributions are then only calculated for the top-$u$ queries under query sparsity measurement. The rest of the queries are assigned with discrete uniform distributions.

\subsubsection{Attention with Compressed Key-Value Memory}\label{sec:compressed_mem}

Apart from decreasing the number of queries with query prototyping, one can also reduce the complexity by reducing the number of the key-value pairs before applying the attention mechanism, as depicted in Fig. \ref{fig:compressed_mem}.

\citet{liu2018generating} propose Memory Compressed Attention (MCA) that reduces the number of keys and values using a strided convolution. This modification is used as a complement to local attention proposed in the same work (as discussed in Sec.~\ref{sec:sparseattn}), in that it can capture global context. The mechanism reduces the number of keys and values by a factor of kernel size $k$ and thus allowing to process significantly longer sequences than vanilla Transformer given the same computation resources.

Set Transformer~\cite{lee2019set} and Luna~\cite{ma2021luna} use a number of external trainable global nodes to summarize information from inputs and then the summarized representations serve as a compressed memory that the inputs attend to. This reduces the quadratic complexity of self-attention to linear complexity w.r.t. sequence length.

Linformer~\cite{wang2020linformer} utilizes linear projections to project keys and values from length $n$ to a smaller length $n_k$. This also reduces the complexity of self-attention to linear. The drawback of this approach is that an input sequence length has to be assumed and hence it cannot be used in autoregressive attention.

Poolingformer~\cite{zhang2021poolingformer} adopts two-level attention that combines a sliding window attention and a compressed memory attention. The compressed memory module is used after the sliding window attention to increase the receptive field. They explore a few different pooling operations as the compression operation to compress the number of keys and values, including max pooling and pooling with Dynamic Convolution~\cite{wu2019paylessattention}.

\subsection{Low-rank Self-Attention}

Some empirical and theoretical analyses~\cite{guo19lowrank,wang2020linformer} report the self-attention matrix $\bA \in \mathbb{R}^{T\times T}$ is often low-rank\footnote{The rank of $\bA$ is far lower than input length $T$.}. The implications of this property are twofold: (1) The low-rank property could be explicitly modeled with parameterization; (2) The self-attention matrix could be replaced by a low-rank approximation.

\subsubsection{Low-rank Parameterization}
The fact that the rank of the attention matrix is less than sequence length implies that, for scenarios where the inputs are typically short, setting $D_k>T$ would be more than an over-parameterization and lead to overfitting. It is thus reasonable to limit the dimension of $D_k$ to explicitly model the low-rank property as an inductive bias. \citet{guo19lowrank} decompose self-attention matrix into a low-rank attention module with small $D_k$ that captures long-range non-local interactions, and a band attention module that captures local dependencies.

\subsubsection{Low-rank Approximation}
Another implication of the low-rank property of the attention matrix is that one can use a low-rank matrix approximation to reduce the complexity of self-attention. A closely related methodology is the low-rank approximation of kernel matrices. We believe some existing works are inspired by kernel approximation.

Some of the aforementioned linearized attention methods in Sec.~\ref{sec:kernel} are inspired from kernel approximation with random feature maps. For example, Performer~\cite{choromanski2020masked} follows the Random Fourier feature map originally proposed to approximate Gaussian kernels. The method first decomposes the attention distribution matrix $\bA$ into $\bC_Q\bG\bC_K$ where $\bG$ is a Gaussian kernel matrix and the random feature map is used to approximate $\bG$.

Another line of work follow the idea of Nystr\"om method. These Nystr\"om-based methods~\cite{chen20compressedlowrank,xiong21nystromformer} first select $m$ landmark nodes from the $T$ inputs with down-sampling methods (e.g., strided average pooling). Let $\tilde\bQ,\tilde\bK$ be the selected landmark queries and keys, then the follow approximation is used in the attention computation
\begin{equation}\label{eq:nystrom}
    \tilde{\bA}=\mathrm{softmax}\left(\bQ\tilde{\bK}^\top\right)\left(\mathrm{softmax}\left(\tilde\bQ\tilde{\bK}^\top\right)\right)^{-1}\mathrm{softmax}\left(\tilde\bQ\bK^\top\right).
\end{equation}

Note that $\bM^{-1}=\left(\mathrm{softmax}\left(\tilde\bQ\tilde{\bK}^\top\right)\right)^{-1}$ in Eq. \eqref{eq:nystrom} does not always exist. To mitigate this issue, CSALR~\cite{chen20compressedlowrank} adds an identity matrix to $\bM$ to make sure that the inverse always exists. Nystr{\"{o}}mformer~\cite{xiong21nystromformer} uses the Moore-Penrose pseudoinverse of $\bM$ instead of the inverse so that the approximation can be made for cases where $\bM$ is singular.

\subsection{Attention with Prior}

\begin{figure}[htbp]
    \centering
    \includegraphics[width=0.5\linewidth]{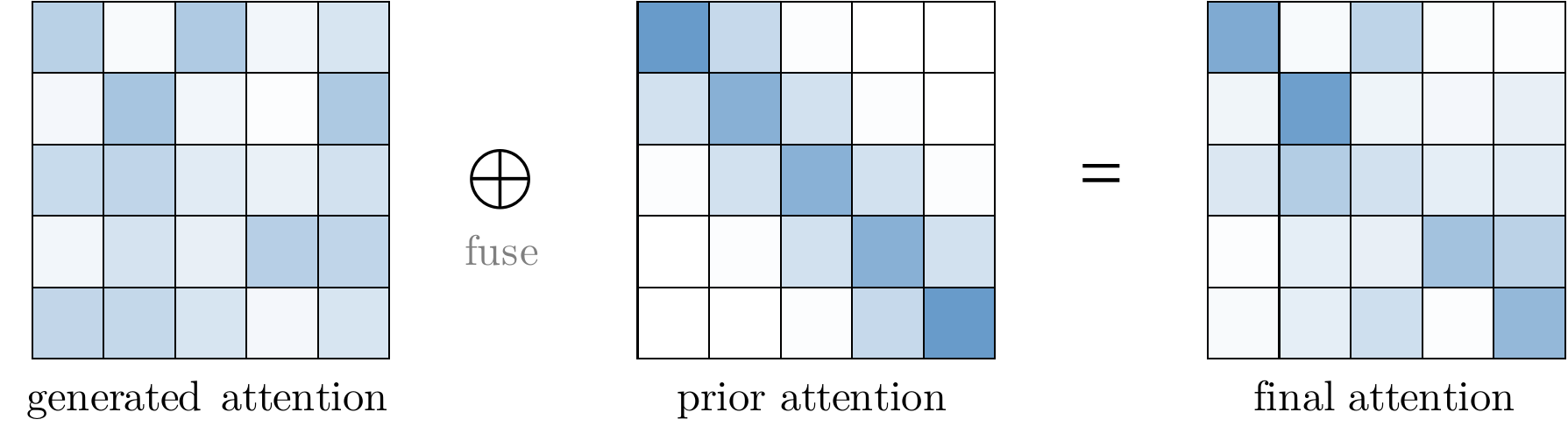}
    \caption{Attention with prior. This type of model fuse generated attention scores with pior attention scores, producing the final attention scores for attention computation.}
    \label{fig:attention_w_prior}
\end{figure}

Attention mechanism generally outputs an expected attended value as a weighted sum of vectors, where the weights are an attention distribution over the values. Traditionally, the  distribution is generated from inputs (e.g., $\mathrm{softmax}(\bQ\bK^\top)$ in vanilla Transformer). As a generalized case, attention distribution can also come from other sources, which we refer to as \textit{prior}. Prior attention distribution can be a supplement or substitute for distribution generated from inputs. We abstract this formulation of attention as \textit{attention with prior}, as depicted in Fig. \ref{fig:attention_w_prior}. In most cases, the fusion of two attention distribution can be done by computing a weighted sum of the scores corresponding to the prior and generated attention before applying softmax.

\subsubsection{Prior that Models locality} Some types of data (e.g., text) can exhibit a strong preference for the locality. This property can be explicitly encoded as a prior attention. A simple method would be to use a Gaussian distribution over positions. Specifically, one could multiply the generated attention distribution with some Gaussian density and then renormalize, which is equivalent to adding to the generated attention scores $\bA$ a bias term $\bG$, where higher $G_{ij}$ indicates a higher prior probability that the $i$-th input attend to the $j$-th input.

\citet{yang18localness} proposes to first predict a central position $p_i$ for each $\bq_i$ using a simple feed-forward network. The Gaussian bias is then defined to be
\begin{equation}
    G_{ij}=-\frac{(j-p_i)^2}{2\sigma^2},
\end{equation}
where $\sigma$ denotes standard deviation for the Gaussian and can be determined as a hyperparameter or predicted from inputs.

Gaussian Transformer~\cite{guo19gaussian} assumes the central position to be $i$ for each $\bq_i$ and defines the bias to bes
\begin{equation}
    G_{ij}=-|w(i-j)^2+b|,
\end{equation}
where $w\ge 0, b\le 0$ are scalar parameters that controls the deviation and reduce the weight for central position, respectively.

\subsubsection{Prior from Lower Modules}\label{sec:prior_prev_module}

In Transformer architecture, it is often observed the attention distributions are similar in adjacent layers. It is thus natural to provide attention distribution from previous layer as a prior for attention computation. The final attention scores can be defined as
\begin{equation}\label{crosslayer}
    \hat{\bA}^{(l)} = w_1\cdot\bA^{(l)}+w_2\cdot g(\bA^{(l-1)}),
\end{equation}
where $\bA^{(l)}$ denotes the attention scores of the $l$-th layer, $w_1,w_2\in\mathbb R$ are weight applied to the scores from adjacent layers, and $g:\mathbb{R}^{n\times n}\rightarrow \mathbb{R}^{n\times n}$ is a function that translate previous scores to the prior to be applied.

Predictive Attention Transformer~\cite{wang2021predictive} proposes to apply a 2D-convolutional layer to previous attention scores and compute the final attention scores as a convex combination of the generated attention scores and the convolved scores. This is equivalent to setting $w_1=\alpha,w_2=1-\alpha$ and $g(\cdot)$ to be a convolutional layer in Eq. \eqref{crosslayer}. They experiment training such a model from scratch and finetune after adapting the pre-trained BERT model, and both sets of experiments show improvements over baseline models.

Realformer~\cite{he2020realformer} uses adds the previous attention scores directly to the generated attention scores, thus resembles a residual skip connection on attention maps. It's equivalent to setting $w_1=w_2=1$ and $g(\cdot)$ to be identity map in Eq. \eqref{crosslayer}. They conduct pre-training experiments on this model. The results show that this model outperforms the baseline BERT model in multiple datasets and surpasses the baseline model even when pre-training budgets are significantly lower.

As an extreme case, Lazyformer~\cite{ying21lazyformer} proposes to share attention maps between a number of adjacent layers. This is equivalent to setting $g(\cdot)$ to identity and switch the settings of $w_1=0,w_2=1$ and $w_1=1,w_2=0$ alternatingly. The benefit of this approach is that the attention maps are computed only once and reused several times in the succeeding layers, thus reducing the computation cost. Their pre-training experiments show that the resulting model remains effective while being much more efficient to compute.

\subsubsection{Prior as Multi-task Adapters}
Adapters are task-dependent, trainale modules that are attached in specific locations of a pre-trained network for cross-task efficient parameter sharing~\cite{rebuffi17residualadapters}. \citet{pilault2021conditionally} propose a Conditionally Adaptive Multi-Task Learning (CAMTL) framework that uses a trainable attention prior $M(\bz_i)$ that depends on task encoding $\bz_i\in\mathbb{R}^{D_z}$
\begin{equation}
    M(\bz_i)=\bigoplus_{j=1}^m A'_j(\bz_i),\quad A'_j(\bz_i)=A_j\gamma_i(\bz_i)+\beta_i(\bz_i),
\end{equation}
where $\bigoplus$ denotes direct sum, $A_j\in \mathbb{R}^{(n/m)\times(n/m)}$ are trainable parameters, and $\gamma_j,\beta_j:\mathbb{R}^{D_z}\rightarrow \mathbb{R}^{(n/m)\times(n/m)}$ are are Feature Wise
Linear Modulation functions~\cite{perez18film}. A maximum sequence length $n_{max}$ is specified in implementation. The prior is formulated as a block diagonal matrix and added to the attention scores of upper layers in pre-trained Transformers to serve as an adapter for parameter-efficient multi-task inductive knowledge transfer.

\subsubsection{Attention with Only Prior}
Some works have explored using an attention distribution that is independent of pair-wise interaction between inputs. In other words, their models exploit only a prior attention distribution.

\citet{zhang-etal-2018-average} design an efficient Transformer decoder variant called average attention network that uses a discrete uniform distribution as the sole source of attention distribution. The values are thus aggregated as a cumulative-average of all values. To improve the expressiveness of the network, they further adds a feed-forward
gating layer on top of the average attention module. The advantage of this approach is that the adapted Transformer decoder can train in a parallel manner as usual Transformers do and decode like an RNN, thus avoiding the $\mathcal O(T^2)$ complexity in decoding.

\citet{you2020hardcoded} utilize a Gaussian distribution as the hardcoded attention distribution for attention calculation. The intuition is very similar to \citet{yang18localness} and \citet{guo19gaussian} in that attention distribution should be focused on a certain local window. Distinctively, they drop the generated attention completely and use only the Gaussian distribution for attention computation. In this approach, the mean (central position) and variance are designed to be hyperparameters. The experiments show that the hardcoded attention, when applied only to self-attention, can achieve comparable performance to the baseline model in machine translation tasks.

Synthesizer~\cite{synthesizer} proposes to replace generated attention scores with: (1) a learnable, randomly initialized attention scores, and (2) attention scores output by a feed-forward network that is only conditioned on the querying input itself. The experiments on machine translation and language modeling show that these variants can achieve competitive performance with vanilla Transformer. It is not explained why these variants work but the empirical results are intriguing.

\subsection{Improved Multi-Head Mechanism}
Multi-head attention is appealing for the ability to jointly attend to information from different representation subspaces at different positions. However, there is no mechanism to guarantee that different attention heads indeed capture
distinct features.

\subsubsection{Head Behavior Modeling}
A basic motivation for using multi-head attention is to allow the model to jointly attend to information from different representation subspaces at different positions~\cite{vaswani2017attention}. However, in vanilla Transformer there is no explicit mechanism to guarantee different behavior across attention heads, nor is there any mechanism for heads to interact with each other. A line of work is dedicated to improving multi-head mechanism by introducing incorporating more sophisticated mechanisms that guide the behavior of different attention heads or allow interaction across attention heads.

\citet{li18disagreementregu} introduce an auxiliary disagreement regularization term into loss function to encourage diversity among different attention heads. Two regularization terms are respectively to maximize cosine distances of the input subspaces and output representations, while the last one is to disperse
the positions attended by multiple heads with element-wise multiplication of the corresponding attention matrices.

Several probing works have revealed that pre-trained Transformer models exhibit certain patterns of self-attention that are of little linguistic backing. As a representative work, \citet{Kovaleva19bertsecret} identify several simple attention patterns in BERT. For instance, many of the attention heads simply pay attention to special BERT tokens \texttt{[CLS]} and \texttt{[SEP]}. As a result, some constraints can be introduced to boost the training of Transformer models. To this end, \citet{deshpande20guidingattention} propose to use an auxiliary loss, which is defined to be the Frobenius norm between attention distribution maps and predefined attention patterns.

Talking-head Attention~\cite{shazeer20talkinghead} uses a talking head mechanism that linearly projects the generated attention scores from $h_k$ to $h$ heads, applies softmax in that space, and then projects to $h_v$ heads for value aggregation. The motivation is to encourage the model to move information between attention heads in a learnable fashion.

Collaborative Multi-head Attention~\cite{cordonnier20collaborate} uses shared query and key projection $\bW^Q$ and $\bW^K$ and a mixing vector $\bm_i$ for the $i$-th head to filter from the projection parameters such that Eq. \eqref{eq:headi} is adapted to
\begin{equation}
    \mathrm{head}_i=\mathrm{Attention}(\bQ\bW^Q\mathrm{diag}(\bm_i),\bK\bW^K, \bV\bW_i^V),
\end{equation}
where $\bW^Q$ and $\bW^K$ are shared by all the attention heads.

\subsubsection{Multi-head with Restricted Spans}

Vanilla attention adopts full attention spans assume, where a query can attend to all of the key-value pairs. However, it is often observed that some heads focus their attention distribution mainly in a local context while some other heads attend to broader contexts. It could thus be beneficial to restrict the attention spans:
\begin{itemize}
    \item \textit{Locality}. Restricting attention spans induce explicit local constraints. This is advantageous in cases where locality is an important prior.
    \item \textit{Efficiency}. If implemented appropriately, such a model can scale to very long sequences without introducing additional memory footprint and computational time.
\end{itemize}

\begin{figure}[htbp]
\begin{center}
\subfigure[mask function for vanilla attention]{
\includegraphics[width=0.3\linewidth]{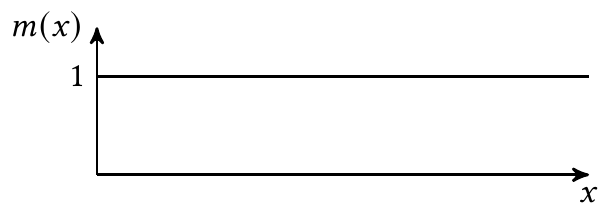}
}
\subfigure[mask function for adaptive span]{
\includegraphics[width=0.3\linewidth]{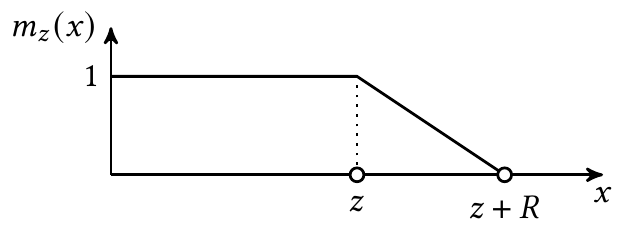}
}
\subfigure[mask function for fixed span]{
\includegraphics[width=0.3\linewidth]{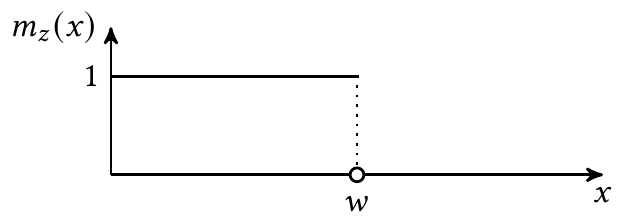}
}
\caption{Three types of span masking function $m(x)$. The horizontal axis represents distance $x$ and vertical axis the mask value.}\label{span_mask}
\end{center}
\end{figure}

Restricting attention spans can be expressed as multiplying each attention distribution value with a mask value and then re-normalize, where the mask can be expressed as a non-increasing function that maps a distance to a value in $[0,1]$. A vanilla attention assigns a mask value of $1$ for all distances, as depicted in Fig. \ref{span_mask}(a).

\citet{Sukhbaatar19adaptivespan} propose to use a learnable attention span, as depicted in Fig. \ref{span_mask}(b) . The mask is parameterized by a learnable scalar $z$ and a hyperparameter $R$. The experiments on character-level language modeling show that the adaptive-span models outperform baseline models while having significantly fewer FLOPS. It is also observed that lower layers generally have smaller learned spans and higher layers otherwise. This indicates that the model can learn a hierarchical composition of features.

Multi-Scale Transformer~\cite{guo2019multiscale} proposes to use a fixed attention span, with different heads in different layers using a different max span. The fixed attention span is depicted in  Fig. \ref{span_mask}(c). The attention is restricted within a fixed window which is controlled by a \textit{scale} value $w$.  They design the scales from an intuitive linguistic perspective and empirical observation from BERT such that higher layers tend to have more large scales (e.g., large span size), and lower layers should be confined with a smaller scale. Their experiments on several tasks show that the model can outperform baseline models while accelerating inference on long sequences.

\subsubsection{Multi-head with Refined Aggregation}

After each attention head computes its output representation, the vanilla multi-head attention~\cite{vaswani2017attention} concatenates these representation and then apply a linear transformation to the concatenated representation to obtain the final output representation, as formulated in Eq. \eqref{eq:multihead}. Combining Eq. \eqref{attention}\eqref{eq:multihead} and \eqref{eq:headi}, one can see that this \textit{concatenate-and-project} formulation is equivalent to summation over $H$ re-parameterized attention outputs. To this end, we first divide $\bW^O\in\mathbb{R}^{D_m\times D_m}$ into $H$ blocks
\begin{equation}
    \bW^O=[\bW_1^O;\bW_2^O;\cdots;\bW_H^O],
\end{equation}
where each $\bW_i^O$ is of dimension $D_v\times D_m$. It's thus easy to see that multi-head attention can be reformulated as
\begin{equation}
    \mathrm{MultiHeadAttn}(Q,K,V)= \sum_{i=1}^H \mathrm{Attention}(Q \bW_i^Q,K \bW_i^K, V\textcolor[rgb]{0,0,1} {\bW_i^V\bW_i^O}).
\end{equation}

One might argue that this simple \textit{aggregate-by-summation} paradigm does not fully exploit the expressiveness of multi-head attention and that it is more desirable to use a more complex aggregation.

\citet{li-etal-2019-multiheadrout,gu19capsule} propose to use routing methods, originally proposed for capsule networks~\cite{sabour17dynamicrouting}, to further aggregate information produced by different attention heads. The outputs of attention heads are first transformed into input capsules, then output capsules are obtained after the iterative routing process. The output capsules are then concatenated as a final output of multi-head attention. These two works both utilizes two routing mechanisms, namely \textit{dynamic routing}\cite{sabour17dynamicrouting} and \textit{EM routing}\cite{hinton18emrouting}. One would notice that iterative routing introduces additional parameters and computational overhead. \citet{li-etal-2019-multiheadrout} empirically show that applying the routing mechanism only to the lower layers can best balance the translation performance and computational efficiency.

\subsubsection{Other Modifications}

Several other modifications to the multi-head mechanism have been proposed to improve multi-head attention.

\citet{shazeer19multiquery} propose multi-query attention, where key-value pairs are shared among attention heads (i.e., to use only one key projection and one value projection for all attention heads). The advantage of this method is that it reduces the memory bandwidth requirements for decoding and results in a model that is faster to decode, while incurring only minor quality degradation from the baseline.

\citet{bhoj20lowrankbottleneck} establish that small attention key size can affect its ability to represent arbitrary distribution. They thus propose to disentangle head size from the number of heads $h$, as opposed to the common practice that sets the head size to be $D_m/h$. It is observed empirically that setting attention head size to be input sequence length is beneficial.

\section{Other Module-level Modifications}\label{sec:other_module}
\subsection{Position Representations}\label{sec:pos_rep}
\begin{definition}[permutation equivariant function]
Let $\Pi_n$ be the set of all permutations of indices $\{1,2,\cdots, T\}$. A function $f:\mathcal{X}^T\rightarrow \mathcal{Y}^T$ is said to be \textit{permutation equivariant} if and only if for any $\pi\in\Pi_T$
\begin{equation}
    f(\pi x)=\pi f(x).
\end{equation}
\end{definition}

It is easy to verify that Convolution and Recurrence networks are not permutation equivariant. However, both self-attention modules and position-wise feed-forward layers in Transformer are permutation equivariant, which could be a problem when it comes to modeling problems other than \textit{set-input} problems where the structure of inputs is needed. For example, when modeling sequences of text, the ordering of words matters and it's thus crucial to properly encode the positions of words in Transformer architecture. Therefore, additional mechanisms are required to inject positional information into Transformers. A common design is to first represent positional information using vectors and then infuse the vectors to the model as an additional input.

\subsubsection{Absolute Position Representations}

In vanilla Transformer~\cite{vaswani2017attention}, positional information is encoded as absolute sinusoidal position encodings.For each position index $t$, the encoding is a vector $\bp_t = \mathrm{PE}(t)\in\mathbb{R}^{D_m}$, of which every element is a sinusoidal ($\sin$/$\cos$) function of the index with pre-defined frequency.
\begin{equation}\label{eq:PE}
    \mathrm{PE}(t)_i = \begin{cases} \sin(\omega_i t) & \text{if }i\text{ is even},\\
    \cos(\omega_i t) & \text{if }i\text{ is odd},
    \end{cases}
\end{equation}
where $\omega_i$ is the hand-crafted frequency for each dimension. The position encoding of each position in the sequence is then added to the token embeddings and fed to Transformer.

Another way of representing absolute positions is to learn a set of positional embeddings for each position~\cite{gehring2017convolutional,devlin2019bert}. Compared to hand-crafted position representation, learned embeddings are more flexible in that position representation can adapt to tasks through back-propagation. But the number of embeddings is limited up to a maximum sequence length determined before training, which makes this approach no longer \textit{inductive}, i.e., not able to handle sequences longer than sequences seen in the training time\cite{liu2020floater,chu2021conditional}.

\citet{wang2021peonbert} propose to use sinusoidal position representation, but with each frequency $\omega_i$ (in Eq. \eqref{eq:PE}) learned from data. This approach retains inductiveness but is more flexible than hand-crafted sinusoidal encoding. FLOATER~\cite{liu2020floater} frames positional representation as a continuous dynamical system and adopts Neural ODE to enable end-to-end training with backpropagation. This method is inductive and flexible while being parameter efficient compared to a fully learnable approach.

The Vanilla approach to incorporating absolute position representations is to add position encodings/embeddings to token embeddings. However, as the input signals propagate through
the layers, the positional information might get lost in the upper layers. Later works find it beneficial to add position representations to inputs to each Transformer layer~\cite{alrfou2018characterlevel,universalxformer,liu2020floater,guo19lowrank}.

\subsubsection{Relative Position Representations}

Another line of works focuses on representing positional relationships between tokens instead of positions of individual tokens. The intuition is that in self-attention, pairwise positional relationships between input elements (direction and distance) could be more beneficial than positions of elements. Methods following this principles are called relative positional representation. \citet{shaw2018relative} propose to add a learnable relative position embedding to keys of attention mechanism
\begin{align}
    \bk_j' &= \bk_j + \br_{ij},\ \text{for }i=1,\cdots, n,\\
    \br_{ij} &= \bR_{\mathrm{clip}(i-j)}\label{rij},\\
    \mathrm{clip}(x) &= \max(-K, \min(x, K))\label{eq:shaw_clip},
\end{align}
where $\br_{ij}\in\mathbb{R}^{D_k}$ is the relative position embedding for relation between position $i$ and $j$ and $K$ is the largest offset that determines the number of embeddingg. Typically $K$ is set to a length that can accommodate most input sequences. As a special case, InDIGO~\cite{gu19indigo} sets $K$ to $3$ for their specially designed framework for non-autoregressive generation. As an incremental effort, Music Transformer~\cite{huang2018music} further introduce a mechanism to reduce the intermediate memory requirements for this approach. Similar to this approach, T5~\citet{raffel2020t5} adopt a simplified form of relative position embeddings where each embedding is only a learnable scalar that is added to the corresponding score used for computing the attention weights.

Transformer-XL~\cite{dai2019transformerxl} use a sinusoidal encoding to represent positional relationships but fuses contents and position information by redesign the computation of attention scores\footnote{the scaling factor is omitted without loss of generality.}
\begin{equation}
    \bA_{ij}=\bq_i \bk_j^\top+\bq_i\left(\bR_{i-j}\bW^{K,R}\right)^\top +\bu^1\bk_j^\top+\bu^2\left(\bR_{i-j}\bW_{K,R}\right)^\top,
\end{equation}
where $\bW^{K,R}\in\mathbb{R}^{D_m\times D_k},\bu^1,\bu^2\in\mathbb{R}^{D_k}$ are learnable parameters and $\bR$ is a sinusoidal encoding matrix similar to position encoding in vanilla Transformer. Then softmax function is applied to scores $\bA$ to provide attention weights. Note that the learnable sinusoidal encoding\cite{wang2021peonbert} is also a drop-in replacement to hand-crafted $\bR$.

DeBERTa~\cite{he2020deberta} utilizes position embeddings like \citet{shaw2018relative} and applies the embeddings to the model in a disentangled style similar to Transformer-XL~\cite{dai2019transformerxl}
\begin{equation}
    \bA_{ij}=\bq_i \bk_j^\top+\bq_i\left(\br_{ij}\bW^{K,R}\right)^\top +\bk_j\left(\br_{ij}\bW^{Q,R}\right)^\top,
\end{equation}
where $\bW^{K,R},\bW^{Q,R}\in\mathbb{R}^{D_m\times D_k}$ are learnable parameters and $\br_{ij}$ is the learnable relative positional embedding as in Eq. \eqref{rij}. The first term is interpreted as a content-to-content attention, and the latter two terms are interpreted as (relative) content-to-position and position-to-content attention, respectively.

\subsubsection{Other Representations}
Some research studies have explored using hybrid positional representations that contains both absolute and relative positional information. Transformer with Untied Position Encoding (TUPE)~\cite{ke2020tupe} re-designs the computation of attention scores as a combination of a content-to-content term, an absolute position-to-position term and a bias term representing relative positional relationships
\begin{equation}
    \bA_{ij}=\bq_i \bk_j^\top+\left(\bp_i\bW^{Q,P}\right)\left(\bp_{j}\bW^{K,P}\right)^\top +b_{j-i},
\end{equation}
where $\bW^{K,P},\bW^{Q,P}\in\mathbb{R}^{D_m\times D_k}$ are learnable parameters, $\bp_i, \bp_j$ are the position embeddings for positions $i,j$, and $b_{j-i}$ is a learnable scalar relative position embedding.

One can also design a single set of positional representations that express both absolute and relative information. Roformer~\cite{su2021roformer} uses Rotary Position Embedding (RoPE) to represent the position of a token by multiplying the affine-transformed embedding of the $t$-th input $x_t$ by a rotatory matrix $\bR_{\Theta,t}$
\begin{align}
    \bq_t=\bx_t\bW^Q\bR_{\Theta,t}\quad & \bk_t=\bx_t\bW^K\bR_{\Theta,t},\\
    \bR_{\Theta,t}&=\bigoplus_{j=1}^{D_k/2} \bM(t,\theta_j),
\end{align}
where $\bigoplus$ denotes \textit{direct sum} of matrices. Each $\bM(t,\theta_j)$ is a 2-D clockwise rotatory matrix of angle $t\cdot \theta_j$
\begin{equation}
    \bM(t,\theta_j)=\left[\begin{matrix} \cos(t\cdot \theta_j)&\sin(t\cdot \theta_j)\\
    -\sin(t\cdot \theta_j)&\cos(t\cdot \theta_j)\end{matrix}
    \right].
\end{equation}

The key advantage of this formulation is that the induced representation is translation invariant, i.e., the attention score of $(\bq_i,\bk_j)$ is only related to their relative position offset
\begin{equation}
    \bq_i\bk_j^\top=\left(\bx_i\bW^Q\right)\bR_{\Theta,\textcolor[rgb]{0,0,1}{j-i}}\left(\bx_j\bW^K\right)^\top.
\end{equation}
In practice, the embedding matrix multiplication can be implemented by two element-wise multiplication for lower memory footprint. The RoPE uses the form of absolute embedding but can capture relative positional relations. This approach is compatible with linearized attention in Sec.~\ref{sec:kernel}.

\subsubsection{Position Representations without Explicit Encoding}

Instead of explicitly introducing additional positional encodings, \citet{wang2020complex} propose to encode positional information in word embeddings, by generalizing embedding to continuous (complex-valued) functions over positions.

R-Transformer~\cite{wang19rxformer} model locality of sequential data with a local RNN. Specifically, inputs to each block of R-Transformer are first fed to a local RNN and then to multi-Head self-attention module. The RNN structure introduces ordering information and captures local dependencies as a complement to self-attention.

Conditional positional encoding (CPE)~\cite{chu2021conditional} generate conditional position encodings at each layer for ViT with a 2-D convolution with zero-paddings. The intuition behind this approach is that convolution networks can implicitly encode absolute positional information with zero-paddings~\cite{islam20cnnpos}.

\subsubsection{Position Representation on Transformer Decoders}

It is worth noticing that masked self-attention is not permutation equivariant~\cite{tsai-etal-2019-dissection}. Thus a model that exploits only the decoder of Transformer has the potential of sensing positional information without incorporating explicit positional representation. This is confirmed by some empirical results on language modeling tasks~\cite{IrieZSN19LM,schlag21fastweight}, where the authors find that removing position encodings even improves performance.

\subsection{Layer Normalization}

Layer Normalization (LN), along with residual connection, is considered as a mechanism to stabilizing training of deep networks (e.g., alleviating ill-posed gradients and model degeneration).
There are some studies that are dedicated to analyzing and improving LN module.

\subsubsection{Placement of Layer Normalization}
\ifx \smv \undefined
\begin{figure}[htbp]
\begin{center}
\subfigure[post-LN]{
\includegraphics[height=2.2in]{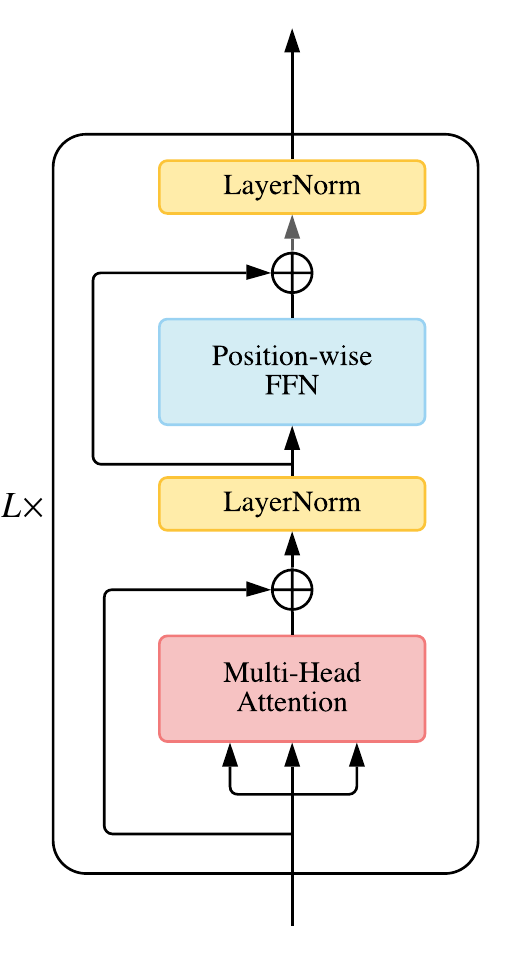}
}\hspace{.6in}
\subfigure[pre-LN]{
\includegraphics[height=2.2in]{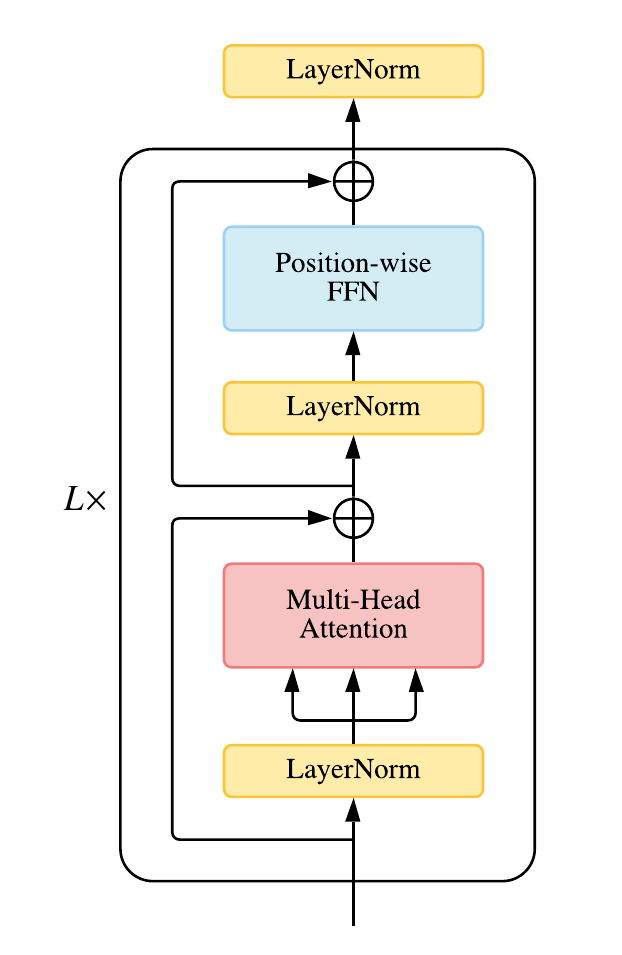}
}
\caption{Comparison of Transformer Encoder with pre-LN and post-LN.}\label{fig:pre_post_ln}
\end{center}
\end{figure}
\fi
In vanilla Transformer, the LN layer lies between the residual blocks, called post-LN~\cite{wang19learningdeep}. Later Transformer implementations~\cite{vaswani-etal-2018-tensor2tensor,klein-etal-2017-opennmt} place the LN layer inside the residual connection before the attention or FFN, with an additional LN after the final layer to control the magnitude of final outputs, which is referred to as pre-LN\footnote{To the best of our knowledge, this approach is adopted since \texttt{v1.1.7} in the \texttt{Tensor2Tensor} implementation~\cite{vaswani-etal-2018-tensor2tensor}.}. The pre-LN has been adopted by numerous following research studies and implementations, e.g., \cite{baevski2019adaptive,child2019generating,wang19learningdeep}.
The difference between pre-LN and post-LN is shown in Fig. \ref{fig:pre_post_ln}.

\citet{xiong2020layer} theoretically investigate the gradients of Transformers and find that the gradients near the output layer are large at initialization in post-LN Transformers, which could be the reason why post-LN Transformers without learning rate warm-up~\cite{vaswani2017attention}\footnote{Learning rate warm-up refers to starting optimization with an extremely small learning rate and then gradually increasing it to a pre-defined maximum value in a certain number of iterations.} leads to unstable training, whereas pre-LN Transformers do not suffer from the same problem. They thus deduce and empirically verify that warm-up stage can be safely removed for pre-LN Transformers.

Although Post-LN often results in unstable training and divergence, it usually outperforms pre-LN variants after convergence~\cite{liu20understanding}. Similar to \citet{xiong2020layer}, \citet{liu20understanding} conduct theoretical and empirical analysis and find that post-LN encoders do not suffer from gradient imbalance. They thus conjecture that the gradient issue is not the direct cause of unstable post-LN Transformer training and further identify the \textit{amplification effect} in post-LN Transformers — at initialization, the heavier dependency on residual branch leads to a larger output shift in post-LN Transformers, thus resulting in unstable training. In light of this finding, they introduce additional parameters to post-LN Transformers to control residual dependencies of Post-LN. These parameters are initialized according to activation variations of sample data so that the output shift of post-LN Transformers is not amplified. This approach ensures and boosts convergence of post-LN Transformers and reaches better performance than pre-LN Transformers.

\subsubsection{Substitutes of Layer Normalization}

\citet{xu19understanding} empirically observe that the learnable parameters in the LN module do not work in most experiments, and even increase the risk of overfitting. They further conclude from controlled experiments that the forward normalization is not the reason why LN works for Transformer. From analysis and experiments, it is concluded that the derivatives of the mean and variance re-center and re-scale the gradients and play a significant role in LN. They thus propose \textit{AdaNorm}, a normalization technique without learnable parameters
\begin{align}
    \bz &= C(1-k\by)\odot\by,\\
    \by &=\frac{\bx-\mu}{\sigma},
\end{align}
where $C,k$ are hyperparameters and $\odot$ denotes element-wise multiplication. $\mu$ and $\sigma$ are the mean and standard deviation of input $\bx$, respectively.

\citet{nguyen19xformerwotears} propose to replace the LN module with \textit{scaled $\ell_2$ normalization}. Given any input $\bx$ of $d$-dimension, their approach project it onto a $d-1$-sphere of learned radius $g$
\begin{equation}
    \bz=g\frac{\bx}{\|\bx\|},
\end{equation}
where $g$ is a learnable  scalar. It is more parameter efficient compared to normal LN and is shown to be effective in machine translation datasets, especially in low-resource settings.

\citet{shen2020powernorm} discuss why Batch Normalization (BN)~\cite{ioffe15batchnorm} performs poorly in Transformer for text data and conclude that BN's significant performance degradation stems from the instabilities associated with its batch statistics. They thus propose PowerNorm (PN) that has three modifications over BN: (1) it relaxes the zero-mean normalization; (2) it uses the quadratic mean
of the signal, instead of the variance; (3) it uses running statistics for the
quadratic mean, instead of using per-batch statistics. Specifically, for the $t$-th iteration, the PN computes the outputs as
\begin{align}
    \bz^{(t)} &= \gamma\odot \by^{(t)}+\beta,\\
    \by^{(t)} &=\frac{\bx^{(t)}}{\psi^{(t-1)}},\\
    (\psi^{(t)})^2&=\alpha(\psi^{(t-1)})^2+(1-\alpha)\left(\frac{1}{|B|}\sum_{i=1}^{|B|}(\bx_i^{(t)})^2\right),
\end{align}
where $0<\alpha<1$ is the moving average coefficient and $\gamma, \beta$ are the learnable parameters as in BN formulation.

\subsubsection{Normalization-free Transformer}

Besides LN, there is another mechanism to construct deeper neural network. ReZero~\cite{bachlechner20rezero} replace LN module with a learnable residual connection. For each module $F(\cdot)$, ReZero re-scales $F(\cdot)$ in the residual formulation:
\begin{equation}
    \bH'=\bH+\alpha\cdot F(\bH),
\end{equation}
where $\alpha$ is a learnable parameter with zero-initialization.

Replacing LN in Transformer with ReZero mechanism is verified to induce better dynamic isometry for input signals and leads to faster convergence.

\subsection{Position-wise FFN}\label{sec:ffn}

Despite its simplicity, the position-wise feed-forward network (FFN) layers are important for a Transformer to achieve good performance. \citet{dong21notallyouneed} observe that simply stacking self-attention modules causes a \textit{rank collapse} problem, leading to token-uniformity inductive bias, and that the feed-forward layer is one of the important building blocks that mitigate this issue.
Various works have explored modifications on the FFN module.

\subsubsection{Activation Function in FFN}

The vanilla Transformer~\cite{vaswani2017attention} adopts the Rectified Linear Units (ReLU) activation for non-linearity in between the two FFN layers. Over time, several studies have explored different activation other than ReLU.

\citet{ramachandran18searching} try to replace ReLU in Transformer with Swish function $f(x)=x\mathrm{sigmoid}(\beta x)$ and observe that it consistently improve performance on WMT 2014 English$\rightarrow$German dataset.

GPT~\cite{radford2018gpt} replace ReLU with Gaussian Error Linear Unit (GELU)~\cite{hendrycks2020gelu} on language pre-training. It becomes the default practice for many pre-trained language models ~\cite{devlin2019bert,he2020deberta}.

\citet{shazeer2020glu} explore using Gated Linear Units (GLU)~\cite{dauphin17glu} and its variants as a drop-in replacement for ReLU in FFN. Their pre-training experiments show that the GLU variants consistently improve vanilla Transformer with ReLU activation. Note that GLU introduces extra parameters and the experiments are conducted with the intermediate dimension of FFN reduced to match the parameter count with baseline.

\subsubsection{Adapting FFN for Larger Capacity}

Several works have focused on expanding FFNs in order for a larger model capacity. The basic idea is to replace FFNs with similar structures with much more parameters.

\citet{lample2019large} replace some of the FFNs with the product-key memory layers. A product-key memory is composed of three components: a query network, a key selection module containing two
sets of sub-keys, and a value lookup table. The model first projects an input to a latent space using the query network, and then compares the generated query to keys that are Cartesian product of the two sets of sub-keys from key selection module to get $k$ nearest neighbors, and finally finds the corresponding values in a value lookup table using the $k$ nearest keys and aggregates them to produce the final output. This process resembles the attention mechanism, in that the generated query attends to a large number of global key-value pairs. They thus propose a multi-head mechanism for the key-product memory to further enlarge the capacity of this module. The experiments on large-scale language modeling suggest that this mechanism significantly improves performance with negligible computational overhead.

Several studies exploits the idea of Mixture-of-Experts (MoE)\cite{shazeer17moe} to increase the capacity of FFNs. Gshard\cite{lepikhin20gshard} uses sparsely-gated MoE layers to replace FFNs in Transformer. Each MoE layer consists of several FFNs (each called an expert) that are the same structure as position-wise FFNs in vanilla Transformer. The output of the layer is a weighted sum of the outputs of the FFNs, using gate values computed by a routing function $g(\cdot)$. They design a learnable routing function that assigns tokens to experts, with auxiliary loss to satisfy balanced loads between experts and efficiency at the scale of length such that the experts can be distributed across multiple devices. For each forward pass of the MoE layer, only the experts with top-$k$ gate values are activated.

Instead of using $k$ experts for each forward pass, Switch Transformer~\cite{fedus2021switch} proposes to route using only a single expert with the largest gate value, leading to a much smaller computational footprint. The authors also design an auxiliary loss to encourage load balance between experts. It is reported to speed up pre-training by a large margin compared to the non-MoE counterpart while having a similar number of FLOPS.

\citet{yang2021exploring} propose to replace top-$k$ routing with expert prototyping strategy. Specifically, the proposed strategy splits experts into $k$ different groups and applies top-1 routing within each group. The outputs of prototype groups are combined linearly to form the final output of the MoE layer. This strategy is proved to improve the model quality while maintaining constant computational costs.

As opposed to using a learnable routing function for expert assignment,  \citet{roller2021hash} design \textit{hash layers} where tokens are hashed into a fixed number of buckets, each bucket corresponding to an expert. This approach requires no routing parameters or any auxiliary loss function, while showing competitive results with existing methods such as Switch Transformer~\cite{fedus2021switch}.

\subsubsection{Dropping FFN Layers}

Notably, one might argue that under some circumstances, FFN layers can be dropped completely, resulting in a simplified network.

\citet{sukhbaatar2019augmenting} demonstrate that replacing the ReLU activation with Softmax and dropping the bias term in FFN effectively turns FFN into an attention module where position-wise inputs attend to a global key-value memory of $D_{\mathrm{ffn}}$ slots. They thus propose to drop the FFN module and add to the attention module a set of global key-value pairs, which are learnable parameters concatenated with key and values generated by inputs. This approach simplifies the structure of the network with no loss of performance.

\citet{yang20xformerdec} empirically show that FFNs in the decoder of Transformer, despite its large number of parameters, is not efficient and can be removed safely with only slight or no loss of performance. This approach significantly boosts the training and inference speed.

\section{Architecture-level Variants}\label{sec:beyond}

In this section, we introduce the X-formers that modify the vanilla Transformer beyond modules.

\subsection{Adapting Transformer to Be Lightweight}

Apart from the efforts made at the module level to alleviate computation overheads, there are several attempts to adapt Transformer to be lightweight by modifications at a higher level.

Similar to low-rank self-attention~\cite{guo19lowrank} that decomposes attention into a locality-constrained attention and a low-rank global attention, Lite Transformer~\cite{wu20longshortrange} proposes to replace each attention module in Transformer with a two-branch structure, where one branch uses attention to capture long-range contexts while the other branch uses depth-wise convolution and linear layers to capture local dependencies. The architecture is lightweight both in terms of model size and computation, and is thus more suitable for mobile devices.

Funnel Transformer~\cite{dai20funneltransformer} utilizes a funnel-like encoder architecture where the length of the hidden sequence is gradually reduced using pooling along the sequence dimension, and then recovered using up-sampling. The architecture effectively reduces the FLOPs and memory compared to the vanilla Transformer encoder. Naturally, one can use this architecture to build a deeper or wider model using the same computation resources.

DeLighT~\cite{mehta2020delight} replaces the standard Transformer block with \texttt{DeLighT} block, which consists of three sub-modules: (1) a ``expand-and-reduce'' \texttt{DeLighT} transformation module to learn wider representations with low computation requirements; (2) a single-head self-attention to learn pair-wise interaction; (3) a lightweight ``reduce-and-expand'' FFN (as opposed to vanilla Transformer that first expands the dimension of hidden representations and then reduces them back to $D_m$). They also propose a block-wise scaling strategy that allows for shallower and narrower blocks near the input and wider and deeper blocks near the output. The induced network is much deeper than the vanilla Transformer but with fewer parameters and operations.

\subsection{Strengthening Cross-Block Connectivity}
In vanilla Transformer, each block takes outputs from the previous block as inputs and outputs a sequence of hidden representations. One might be interested in creating more paths along which input signals can run through the networks. In Sec.~\ref{sec:prior_prev_module}, we introduced Realformer~\cite{he2020realformer} and Predictive Attention Transformer~\cite{wang2021predictive} that reuses attention distributions from previous block to guide attention of current block. This can be seen as creating a forward path between adjacent Transformer blocks.

In a deep Transformer encoder-decoder model, the cross-attention modules in the decoder only utilize the final outputs of the encoder, therefore the error signal will have to traverse along the depth of the encoder.  This makes Transformer more susceptible to optimization issues (e.g., vanishing gradients). Transparent Attention~\cite{Chen18transparentxformer} uses a weighted sum of encoder representations at all encoder layers (including the embedding layer) in each cross-attention module. For the $j$-th decoder block, the cross-attention module is modified to attend to
\begin{equation}\label{eq:transparent}
    \tilde{\bH}^{(j)}=\sum_{i=0}^{N}\frac{\exp(w_{ij})}{\sum_{k=0}^N \exp(w_{kj})}\bH^{(i)},
\end{equation}
where each $w_{ij}$ is a trainable parameter. This effectively shortens the path from each layer in the encoder to the error signal and thus eases the optimization of deeper Transformer models.

Another issue associated with vanilla Transformer is that each position can only attend to history representations from lower layers. Feedback Transformer~\cite{fan2021feedbackmem} proposes to add a feedback mechanism to Transformer decoder, where each position attends to a weighted sum of history representations from all layers
\begin{equation}
    \tilde{\bh}_i=\sum_{l=0}^{N}\frac{\exp(w_{l})}{\sum_{k=0}^N \exp(w_{k})}\bh_i^{(l)}.
\end{equation}

 \subsection{Adaptive Computation Time}

Vanilla Transformer, like most neural models, utilizes a fixed (learned) computation procedure to process each input. An intriguing and promising modification is to make computation time conditioned on the inputs, i.e., to introduce Adaptive Computation Time (ACT)~\cite{graves16act} into Transformer models. Such modifications potentially give rise to the following advantages:
\begin{itemize}
    \item Feature refinement for hard examples. For data that are hard to process, a shallow representation might not be adequate to fulfill the task at hand. It would be more ideal to apply more computations to acquire a deeper and more refined representation.
    \item Efficiency for easy examples. When processing easy examples, a shallow representation might be enough for the task. In this case, it would be beneficial if the network can learn to extract features using reduced computation time.
\end{itemize}

\begin{figure}[htbp]
\begin{center}
\subfigure[dynamic halting]{\label{fig:ut}\hspace{2em}
\includegraphics[height=1.2in]{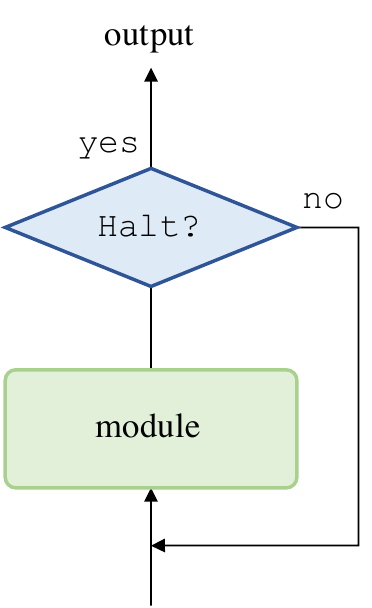}\hspace{2em}
}
\subfigure[conditional skipping]{\label{fig:skip}\hspace{2em}
\includegraphics[height=1.2in]{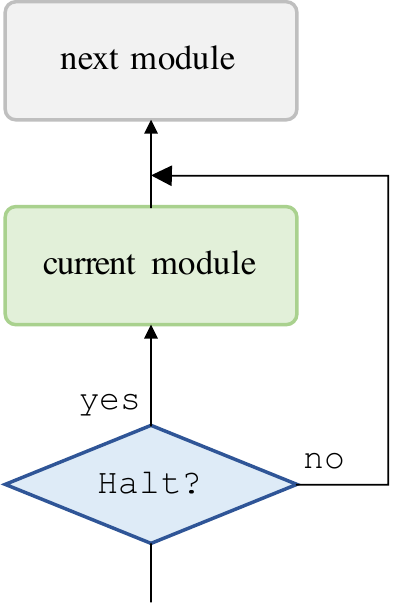}\hspace{2em}
}
\subfigure[early exit]{\label{fig:exit}\hspace{2em}
\includegraphics[height=1.2in]{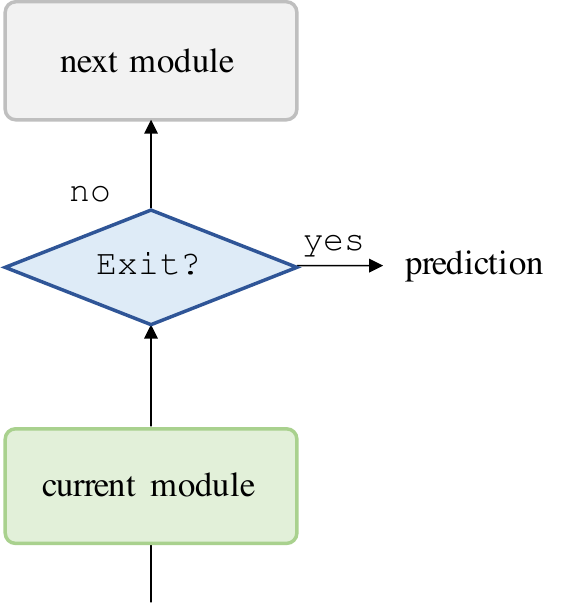}
}
\caption{Three typical ACT paradigms.}\label{fig:act}
\end{center}
\end{figure}

Universal Transformer (UT)~\cite{universalxformer} incorporates a recurrence-over-depth mechanism that iteratively refines representations for all symbols using a module that is shared over depth, as illustrated in Fig. \ref{fig:ut}. It also adds a per-position dynamic halting mechanism that calculates a halting probability for each symbol at every time step. If a symbol's halting probability is greater than a predefined threshold, then the symbol's representation will remain unchanged for subsequent timesteps. The recurrence is stopped when all symbols halt or when a predefined maximum step is reached.

Conditional Computation Transformer (CCT)~\cite{CCTransformer} adds a gating module at each self-attention and feed-forward layer to decide whether to skip the current layer, as illustrated in Fig. \ref{fig:skip}. The authors also introduce an auxiliary loss that encourages the model to adjust the gating modules to match the practical computation cost to the available computation budget.

Similar to the dynamic halting mechanism used in UT, there is a line of work dedicated to adapting the number of layers to each input in order to achieve a good speed-accuracy trade-off, which is called \textit{early exit} mechanism, as illustrated in Fig. \ref{fig:exit}. A commonly used technique is to add an internal classifier at each layer and jointly train all classifiers. The core of these methods is the criteria used to decide whether to exit at each layer. DeeBERT~\cite{xin-etal-2020-deebert} uses the entropy of the output probability distribution of the current layer to determine whether to exit. PABEE~\cite{zhou2020bert} counts the number of times that the predictions remain unchanged to decide whether to exit. \citet{li21accelerating} design a window-based uncertainty criterion to achieve token-level partial exiting for sequence labeling tasks. \citet{sun2021early} introduces a voting-based exiting strategy that considers at each layer predictions of all the past internal classifiers to infer the correct label and to decide whether to exit.

\subsection{Transformers with Divide-and-Conquer Strategies}
\begin{figure}[htbp]
\begin{center}
\subfigure[Recurrent Transformer\label{fig:recurrence}]{
\includegraphics[width=0.4\linewidth]{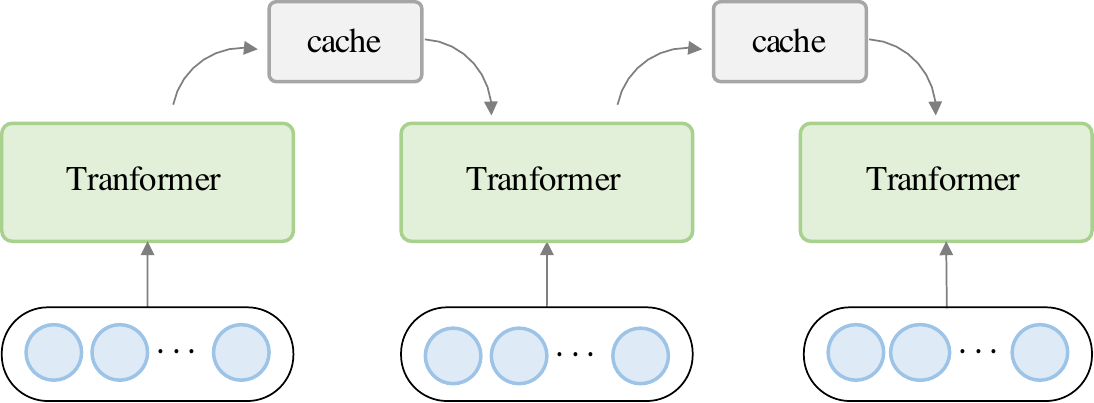}
}\quad
\subfigure[Hierarchical Transformer\label{fig:hierarchy}]{
\includegraphics[width=0.4\linewidth]{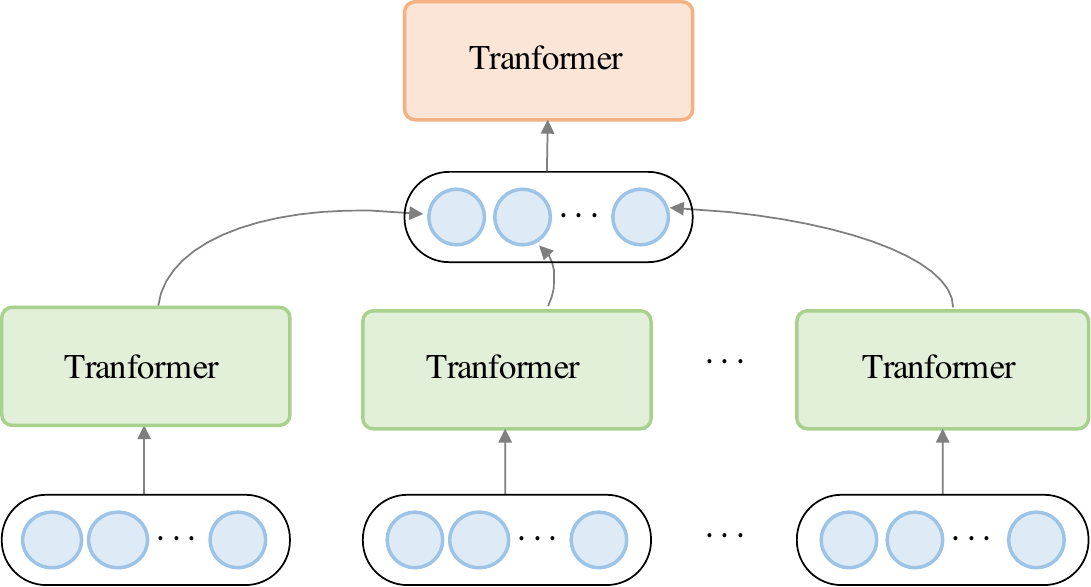}
}
\caption{Illustrations of recurrent and hierarchical Transformers.}\label{fig:divide_and_conquer}
\end{center}
\end{figure}

The quadratic complexity of self-attention on sequences length can significantly limit the performance
of some downstream tasks. For example, language modeling usually needs long-range context.  Apart from the techniques introduced in Sec.~\ref{sec:attention}, another effective way of dealing with long sequences is to use \textit{divide-and-conquer} strategy, i.e., to decompose an input sequence into finer segments that can be efficiently processed by Transformer or Transformer modules. We identify two representative class of methods, \textit{recurrent} and \textit{hierarchical} Transformers, as illustrated in Fig. \ref{fig:divide_and_conquer}. These techniques can be understood as a wrapper for the Transformer model in which Transformer acts as an elementary component that is reused to process different input segments.

\subsubsection{Recurrent Transformers}

In recurrent Transformers, a cache memory is maintained to incorporate the history information. While processing a segment of text, the network reads from the cache as an additional input. After the processing is done, the network writes to the memory by simply copying hidden states or using more complex mechanisms. The abstract process is illustrated in Fig. \ref{fig:recurrence}.

Transformer-XL~\cite{dai2019transformerxl} address the limitation of a fixed length context by caching representations from the previous segment and reuse it as an extended context when the model processes the current segment. For the $l$-th layer and the $(\tau+1)$-th segment, the input representation $\bH_{\tau+1}^{(l-1)}$ is concatenated with the representation $\bH_\tau^{(l-1)}$ from previous segment to produce the keys and values
\begin{align}
    \tilde{\bH}_{\tau+1}^{(l)}&=[\mathrm{SG}(\bH_\tau^{(l-1)})\circ \bH_{\tau+1}^{(l-1)}]\label{eq:xformerxl},\\
    \bK_{\tau+1}^{(l)}, \bV_{\tau+1}^{(l)}&=\tilde{\bH}_{\tau+1}^{(l)}\bW^K, \tilde{\bH}_{\tau+1}^{(l)}\bW^V,
\end{align}
where $\bH_\tau^{(0)}$ is defined as the word embedding sequence, $\mathrm{SG}(\cdot)$ denotes stop-gradient operation and $[\bX\circ\bY]$ denotes concatenating the two vector sequences along the time dimension. This approach extends the maximum context length by $L\times N_{\text{mem}}$ where $L$ is the number of layers and $N_{\text{mem}}$ is the length of cached memory sequence.

Compressive Transformer~\cite{rae2019compressive} extends this idea further by extending the cache with two levels of memory. In Transformer-XL, the activations from the previous segment are cached as a memory that is used to augment the current segment, and activations from older segments are discarded. Compressive Transformer, on the other hand,  applies a compression operation (e.g., Convolution, Pooling, etc.) on older activations and stores them in the compressed memory. In order to avoid the expensive backpropagating-through-time (BPTT) from training compression sub-network with gradients from the loss, they propose to use local loss functions where original memories are constructed from the compressed memories. This approach further extends the theoretical maximum history context length from $L\times N_{\text{mem}}$ of Transformer-XL to $L\times (N_{\text{mem}}+c\times N_{\text{cm}})$, where $c$ is the compression rate and $N_{\text{cm}}$ is the length of compressed memory.

 Memformer~\cite{wu2020memformer} extends the recurrence mechanism from decoder-only architecture to an encoder-decoder architecture. They introduce to the encoder a memory cross attention similar to the cross attention in vanilla Transformer to allow the Transformer encoder to attend to the memory. They also introduce a memory slot attention on top of the encoder output to explicitly write the memory for the next segment. To avoid BPTT over a long range of timesteps, they propose Memory
Replay Back-Propagation (MRBP) algorithm, which replays
the memory at each timestep to accomplish gradient back-propagation over long unrolls.

\citet{yoshida20recurrenceptm} propose a simple fine-tuning mechanism to add recurrence to a pre-trained language model (e.g., GPT-2~\cite{radford2019gpt2}).
\ifx \smv \undefined
They first compress the representations produced by the $\tau$-th segment into one single vector representation, using a weighted average of pooled representations from each layer $l\in\{1,\cdots,L\}$
\begin{equation}
    \bz_\tau=\sum_{l=1}^Lw_l\sum_{j=1}^{T_\tau}\bh_j^{(l)},
\end{equation}
where $T_\tau$ denotes the sequence length of the $\tau$-th segment, $w_l=\mathrm{softmax}(\mathbf\alpha)_l$ is the weight softmax-normalized from learnable parameters $\mathbf\alpha=[\alpha_1,\cdots,\alpha_L]$. This compressed representation is then fed to a  feed-forward network to produce the memory state $\bh_{\text{prev},\tau}$ for the $\tau$-th segment, which is then prepended to the key-value inputs of a specific attention layer. This approach effectively extends the context length of a pre-trained language model, without significant change of the architecture of the original model.
\fi

ERNIE-Doc~\cite{ding2020erniedoc} proposes an enhanced recurrence mechanism based on the recurrence mechanism used in Transformer-XL, by replacing the memory with the history representations from the $l$-th layer.
\ifx \smv \undefined
\begin{align}
    \tilde{\bH}_{\tau+1}^{(l)}&=[\mathrm{SG}(\textcolor[rgb]{0,0,1}{\bH_\tau^{(l)}})\circ \bH_{\tau+1}^{(l-1)}],
\end{align}
as opposed to using representations from the $(l-1)$-th layer in Eq. \eqref{eq:xformerxl}. This modification essentially leads to a larger effective context length.
\fi

\subsubsection{Hierarchical Transformers}
Hierarchical Transformer decomposes inputs hierarchically into elements of finer granularity. Low-level features are first fed to a Transformer encoder, producing output representations that are then aggregated (using pooling or other operations) to form a high-level feature, which is then processed by a high-level Transformer. This class of methods can be understood as a process of hierarchical abstraction. The overview of this approach is depicted in Fig. \ref{fig:hierarchy}. The advantages of this approach are twofold: (1) Hierarchical modeling allows the model to handle long inputs with limited resources; (2) It has the potential to generate richer representations that are beneficial to tasks.

\paragraph{6.5.2.1 Hierarchical for long sequence inputs}
For tasks with inherently long input length, one can use hierarchical Transformers for effective modeling of long-range dependencies. For document-level machine translation tasks, \citet{miculicich-etal-2018-document} introduce dependencies on the previous sentences from both the source and target sides when translating a sentence. They use an attention mechanism as the aggregation operation to summarize low-level information. For document summarization, HIBERT~\cite{zhang-etal-2019-hibert} encodes a document of text by first learn sentence representations for all sentences and then use these sentence representations to encode document-level representations that are then used to generate the summary. The model uses the last hidden representation (corresponding to the \texttt{EOS} token) as the representation for each sentence. \citet{liu-lapata-2019-hierarchical} propose a similar hierarchical Transformer for multi-document summarization where the extracted low-level representations are aggregated using an attention layer with a global trainable query node and low-level representations as the source of key-value pairs. Hi-Transformer~\cite{wu2021hitransformer} first utilizes a sentence Transformer and a document Transformer to hierarchically learn document context-aware sentence representations. The document context-aware sentence representations are then fed to another sentence Transformer to further improve the sentence context modeling.

\paragraph{6.5.2.2 Hierarchical for richer representations}
One might also be interested in using hierarchical models to acquire richer representations that are beneficial to the tasks at hand. For example, TENER~\cite{yan2019tener} uses a low-level Transformer encoder to encode character features, which is then concatenated with word embeddings as the inputs to the high-level Transformer encoder. This incorporates more features and alleviates the problems of data sparsity and out-of-vocabulary (OOV). Recently emerging Vision Transformer~\cite{dosovitskiy2020vit} divides an input image into several patches that serve as the basic input elements of Transformer, which potentially loses intrinsic pixel-level information within patches. To address this issue, Transformer in Transformer (TNT)~\cite{han2021tnt} uses at each layer an inner Transformer block that transforms pixel representations and an outer Transformer block that takes fused vectors of patch representations and pixel representations as input.

\subsection{Exploring Alternative Architecture}
Despite the success of Transformer architecture, one might question whether the current Transformer architecture is optimal. Interestingly, several studies have explored alternative architectures for Transformer.

\citet{lu19macaron} interpret Transformer as a numerical Ordinary Differential Equation (ODE) solver for a convection-diffusion equation in a multi-particle dynamic system and design Macaron Transformer, which replaces each Transformer block with a \textit{FFN-attention-FFN} variant.

Sandwich Transformer~\cite{ofir20sandwichxformer} explores reorganizing attention modules and FFN modules such that attention modules are mainly located in lower layers and FFN modules in upper layers. The induced model improves perplexity on multiple language modeling benchmarks, without increasing parameters, memory or training time.

Mask Attention Network (MAN)~\cite{fan-etal-2021-mask} prepends a dynamic mask attention module to the self-attention module in each Transformer block. The mask is conditioned on token representations, the relative distance between tokens and head indices. The proposed dynamic mask attention is shown to effectively model locality in text data and the induced model consistently outperforms the baseline model in machine translation and abstractive summarization.

Notably, there's a line of work that uses Neural Architecture Search (NAS) to search for alternative Transformer architectures. The Evolved Transformer (ET)~\cite{so19evolvedxformer} employs evolution-based architecture search with the standard  Transformer architecture seeding the initial population. The searched model demonstrates consistent improvement
over Transformer on several language tasks. As another representative work, DARTSformer\cite{zhao2021dartsformer} applies differentiable architecture search (DARTS)~\cite{liu2018darts}, combined with a multi-split reversible network and a backpropagation-with-reconstruction algorithm for memory efficiency. The resulting model consistently outperforms standard Transformer and compares favorably to larger ET models, with a significantly reduced search cost.

\section{Pre-trained Transformers}\label{sec:ptm}
As a key difference from convolutional networks and recurrent networks that inherently incorporates the inductive bias of locality, Transformer does not make any assumption about how the data is structured. On the one hand, this effectively makes Transformer a very universal architecture that has the potential of capturing dependencies of different ranges. On the other hand, this makes Transformer prone to overfitting when the data is limited. One way to alleviate this issue is to introduce inductive bias into the model.

Recent studies suggest that Transformer models that are pre-trained on large corpora can learn universal language representations that are beneficial for downstream tasks~\cite{qiu2020ptms}. The models are pre-trained using various self-supervised objectives, e.g., predicting a masked word given its context. After pre-training a model, one can simply fine-tune it on downstream datasets, instead of training a model from scratch. To illustrate typical ways of using Transformers in pre-training, we identify some of the pre-trained Transformers and categorize them as follows.

\begin{itemize}
    \item \textit{Encoder only}. A line of work uses the Transformer encoder as its backbone architecture. BERT~\cite{devlin2019bert} is a representative PTM that is typically used for natural language understanding tasks. It utilizes Masked Language Modeling (MLM) and Next Sentence Prediction (NSP) as the self-supervised training objective. RoBERTa~\cite{liu2019roberta} further adapts the training of BERT and removes the NSP objective as it is found to hurt performance on downstream tasks.
    \item \textit{Decoder only}. Several studies focus on pre-training Transformer decoders on language modeling. For example, the Generative Pre-trained Transformer (GPT) series (i.e., GPT~\cite{radford2018gpt}, GPT-2~\cite{radford2019gpt2}, and GPT-3~\cite{brown20gpt3}) is dedicated to scaling pre-trained Transformer decoders and has recently illustrated that a large-scale PTM can achieve impressive few-shot performance with the task and examples fed to the model as constructed prompts~\cite{brown20gpt3}.
    \item \textit{Encoder-Decoder}. There are also PTMs that adopt Transformer encoder-decoder as the overall architecture. BART~\cite{lewis20bart} extends the denoising objective of BERT to encoder-decoder architecture. The benefit of using an encoder-decoder architecture is that the inducing model is equipped with the ability to perform both natural language understanding and generation. T5~\cite{raffel2020t5} adopts similar architecture and was one of the earliest studies that use task-specific text prefix in downstream tasks.
\end{itemize}

Some of the Transformer architecture variants can also be applied to Transformer-based PTMs. For instance, BigBird~\cite{zaheer2020big} introduced in Sec.~\ref{sec:sparseattn} is a encoder-based PTM that uses compound position-based sparse attention to enable long sequence inputs. GPT-3~\cite{brown20gpt3} uses alternating dense and locally banded sparse attention (which was also introduced in Sec.~\ref{sec:sparseattn}) in self-attention modules. Switch Transformer~\cite{fedus2021switch} is an encoder-based PTM that replaces FFN layers with mixture-of-experts layers and can increase parameter count while keeping the FLOPs per example constant.

\section{Applications of Transformer}\label{sec:app}
Transformer was originally designed for machine translation but has been widely adopted in various fields besides NLP, including CV and audio processing, due to its flexible architecture.

(1) \textit{Natural Language Processing}. Transformer and its variants have been extensively explored and applied in NLP tasks, e.g., machine translation~\cite{vaswani2017attention,mehta2020delight,raffel2020t5,so19evolvedxformer,fan-etal-2021-mask}, language modeling~\cite{dai2019transformerxl,shoeybi2020megatronlm,roy2020efficient,rae2019compressive} and named entity recognition~\cite{yan2019tener,li-etal-2020-flat}. Massive effort has been dedicated to pre-training Transformer models on large-scale text corpora, which we believe is one of the major reasons of Transformer's wide application in NLP.

(2) \textit{Computer Vision}. Transformer have also been adapted for various vision tasks, e.g., image classification~\cite{chen20igpt,dosovitskiy2020vit,liu2021swin}, object detection~\cite{carion20detr,zhu20deformabledetr,zheng20adaptiveclustering,liu2021swin}, image generation~\cite{parmar2018imagexformer,jiang2021transgan} and video processing~\cite{Shao_2021_WACV,arnab2021vivit}. \citet{han2021surveyvisualxformer} and \citet{khan2021xformersinvision} provide reviews on existing work of visual Transformers. We encourage readers to refer to these surveys for further understand the current research progress on Transformers in CV.

(3) \textit{Audio Applications}. Transformer can also be extended for audio-related applications, e.g., speech recognition~\cite{dong18speechxformer,pham19deepsanspeech,chen20streamingxformer,gulati20conformer}, speech synthesis~\cite{li19xformertts,zheng20rnnenhancedxformer,ihm20reformertts}, speech enhancement~\cite{kim20tgsa,yu21se-xformer} and music generation~\cite{huang2018music}.

(4) \textit{Multimodal Applications}. Owing to its flexible architecture, Transformer has also been applied in various multimodal scenarios, e.g., visual question answering~\cite{li2019visualbert,Hu_2020_CVPR,su20vlbert,li2020unimo}, visual commonsense reasoning~\cite{li2019visualbert,su20vlbert}, caption generation~\cite{sun19videobert,cornia20meshedmemxformer,lin2021m6}, speech-to-text translation~\cite{han2021chimera} and text-to-image generation~\cite{ramesh2021dalle,lin2021m6,ding2021cogview}.

\section{Conclusion and Future Directions}\label{sec:discussion}

In this survey, we conduct a comprehensive overview of X-formers and propose a new taxonomy. Most of the existing works improve Transformer from different perspectives, such as efficiency, generalization, and applications. The improvements include incorporating structural prior, designing lightweight architecture, pre-training, and so on.

Although X-formers have proven their power for various tasks, challenges still exist. Besides the current concerns (e.g. efficiency and generalization), the further improvements of Transformer may lie in the following directions:

(1) \textit{Theoretical Analysis}. The architecture of Transformer has been demonstrated to be capable of supporting large-scale training datasets with enough parameters. Many works show that Transformer has a larger capacity than CNNs and RNNs and hence has the ability to handle a huge amount of training data.
When Transformer is trained on sufficient data,  it usually has better performances than CNNs or RNNs.
An intuitive explanation is that Transformer has few prior assumptions on the data structure and therefore is more flexible than CNNs and RNNs. However, the theoretical reason is unclear and we need some theoretical analysis of Transformer ability.

(2) \textit{Better Global Interaction Mechanism beyond Attention}. A main advantage of Transformer is the use of the attention mechanism to model the global dependencies among nodes within input data. However, many studies have shown that full attention is unnecessary for most nodes. It is, to some degree, inefficient to indistinguishably calculate attention for all nodes. Therefore, there is still plenty of room for improvements in efficiently modeling global interactions. On the one hand, the self-attention module can be regarded as a fully-connected neural network with dynamical connection weights, which aggregates non-local information with dynamic routing. Therefore, other dynamic routing mechanisms are alternative approaches worth exploring. On the other hand, the global interaction can also be modeled by other types of neural networks, such as memory-enhanced models.

(3) \textit{Unified Framework for Multimodal Data}. In many application scenarios, integrating multimodal data is useful and necessary to boost the task performance. Moreover, the general AI also needs the ability to capture the semantic relations across different modalities.
Since Transformer achieves great success on text, image, video, and audio, we have a chance to build a unified framework and better capture the inherent connections among multimodal data. However, the design of the intra-modal and cross-modal attention still remains to be improved.

Finally, we wish this survey to be a hands-on reference for better understanding the current research progress on Transformers and help readers to further improve Transformers for various applications.

\bibliographystyle{ACM-Reference-Format}
\bibliography{references}
\end{document}